\documentclass[11pt]{article}

\usepackage[preprint]{acl}

\usepackage{times}
\usepackage{latexsym}

\usepackage[T1]{fontenc}

\usepackage[utf8]{inputenc}

\usepackage{microtype}

\usepackage{inconsolata}

\usepackage{graphicx}
\usepackage{subcaption}
\usepackage{enumitem}
\usepackage{algorithm}
\usepackage{algorithmicx}
\usepackage{algpseudocode}
\usepackage{amsmath} 
\usepackage{longtable}
\usepackage{multirow}
\usepackage{makecell}
\usepackage{soul}
\usepackage{booktabs}
\usepackage{xcolor}
\usepackage{colortbl}
\usepackage{tcolorbox}      
\usepackage{listings}
\usepackage{amssymb} 
\definecolor{deepgreen}{RGB}{0,180,0}
\lstdefinestyle{python}{
    language=Python,
    basicstyle=\ttfamily\small,
    keywordstyle=\color{blue}\bfseries,
    commentstyle=\color{deepgreen},
    stringstyle=\color{red},
    numberstyle=\tiny\color{gray},
    showstringspaces=false,
    frame=single,
    breaklines=true,
    backgroundcolor=\color{lightgray!15}
}
\newcommand{\myicon}[1]{%
    \raisebox{-0.25ex}{\includegraphics[width=1.0em]{#1}}%
}
\definecolor{myred}{RGB}{255,191,192}

\newcommand{\ourmodel}{\texttt{ET-Agent}}

\title{\myicon{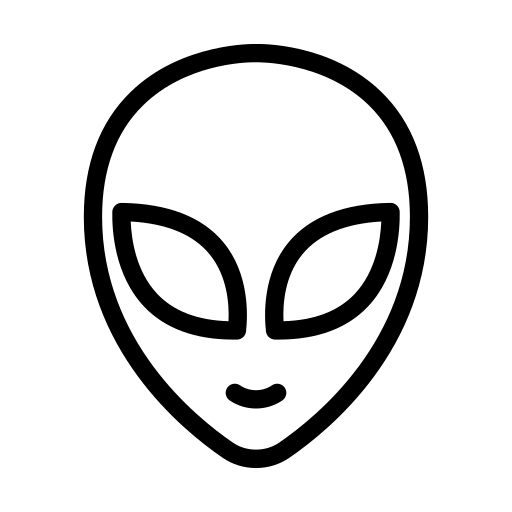}ET-Agent: Incentivizing Effective Tool-Integrated Reasoning Agent via Behavior Calibration} 

\author{Yifei Chen, Guanting Dong, Zhicheng Dou$^{\dag}$ \\
Gaoling School of Artificial Intelligence, Renmin University of China \\
\texttt{zhangboguodong@ruc.edu.cn, dou@ruc.edu.cn}}

\begin{document}
\maketitle
\begin{abstract}
Large Language Models (LLMs) can extend their parameter knowledge limits by adopting the Tool-Integrated Reasoning (TIR) paradigm. However, existing LLM-based agent training framework often focuses on answers' accuracy, overlooking specific alignment for behavior patterns. Consequently, agent often exhibits ineffective actions during TIR tasks, such as redundant and insufficient tool calls. How to calibrate erroneous behavioral patterns when executing TIR tasks, thereby exploring effective trajectories, remains an open-ended problem. In this paper, we propose \ourmodel{}, a training framework for calibrating agent's tool-use behavior through two synergistic perspectives: Self-evolving Data Flywheel and Behavior Calibration Training. Specifically, we introduce a self-evolutionary data flywheel to generate enhanced data, used to fine-tune LLM to improve its exploration ability. Based on this, we implement an two-phases behavior-calibration training framework. It is designed to progressively calibrate erroneous behavioral patterns to optimal behaviors. Further in-depth experiments confirm the superiority of \ourmodel{} across multiple dimensions, including correctness, efficiency, reasoning conciseness, and tool execution accuracy. Our \ourmodel{} framework provides practical insights for research in the TIR field. Codes can be found in ~\url{https://github.com/asilverlight/ET-Agent}
\end{abstract}

\section{Introduction}

Recently, Large Language Models (LLMs) have demonstrated excellent performance in various complex reasoning tasks~\cite{Towards-Reasoning-in-Large-Language-Models:--Survey, deepseekmath,logic-rl,mugglemath,reasoningasaresource,seed18,qiao2024we,qiao2025we}. However, when faced with problems beyond their own knowledge and capabilities, LLMs often struggle to perform effectively~\cite{imitateexploreandselfimprove,o1replicationjourney,programofthoughtsprompting}. As an emerging paradigm, Tool-Integrated Reasoning (TIR) has recently received increasing attention~\cite{tora,start,torl}. LLM-based agents with TIR paradigm can achieve better performance than ordinary reasoning models in tasks such as mathematical reasoning, in-depth search, and scientific question-answering~\cite{torl,search-o1,dong2025toolstarempoweringllmbrainedmultitool}.

\begin{figure}[t]
  \includegraphics[width=\columnwidth]{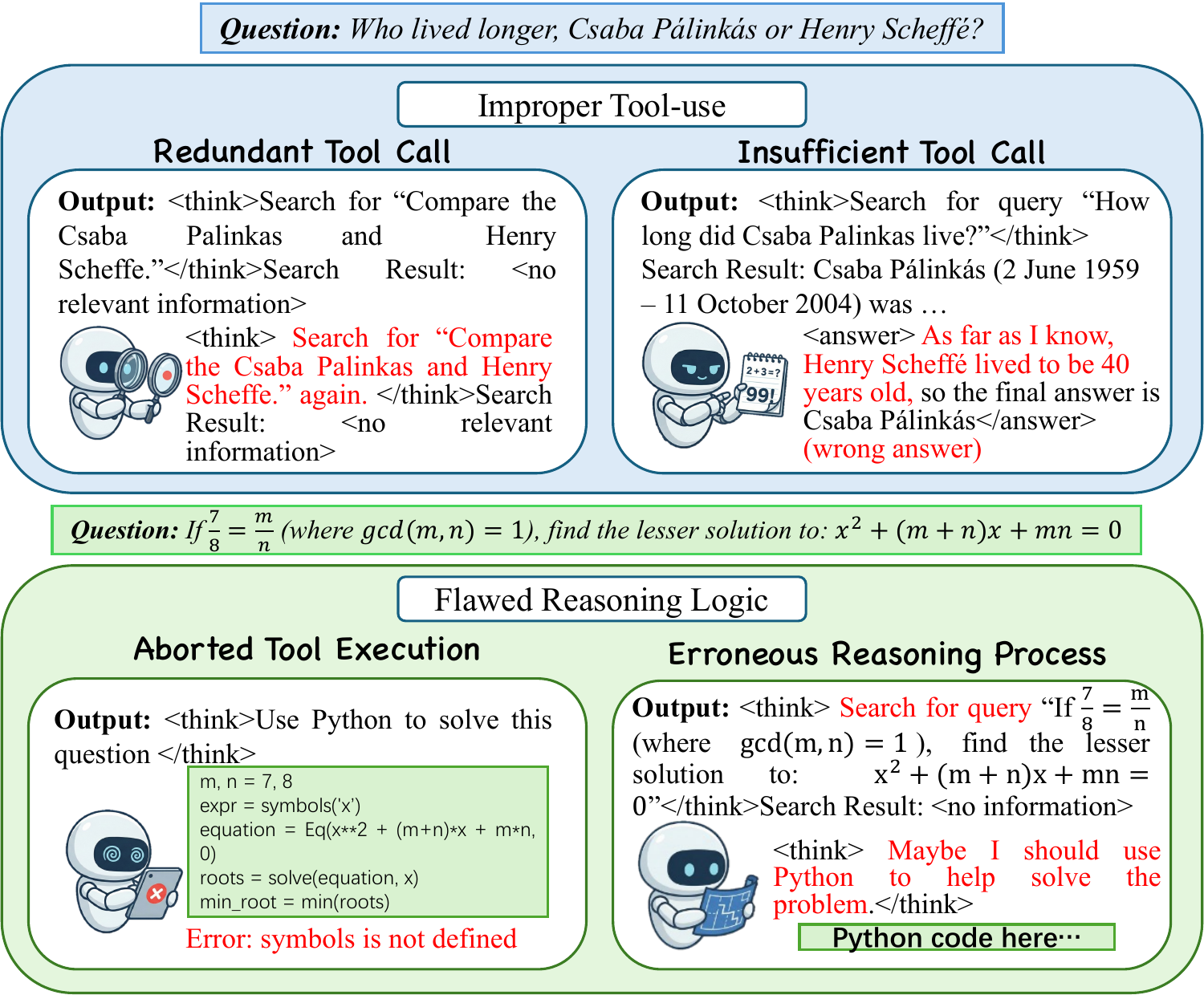}
  \caption{Illustration of two different error tool-use behavior patterns.}
  \label{fig:intro}
\end{figure}

Nevertheless, most current works only focus on the accuracy of TIR paradigm in downstream tasks. There is a lack of necessary attention to TIR's behavioral patterns correctness. For example, some works excessively interact with retrieval systems to achieve deeper and more accurate information acquisition~\cite{beyond10turns,chen2025iterresearchrethinkinglonghorizonagents,webdancer,websailor}. But these methods undoubtedly introduce redundant tool calls, making it more difficult to practically apply these methods.

To optimize TIR behavioral patterns, several works elaborately design data construction processes, aiming to calibrate agents' behaviors to be more efficient~\cite{CPT,smart,dotamath,websailor,tora}. Another branch of efforts start with Reinforcement Learning (RL) algorithms to stimulate LLM's potential to effectively utilize tools~\cite{ikea,self-dc,search-wisely,otc,webleaper}. However, these works only focus on mitigating redundant tool calls, which is very limited. In real TIR scenarios, there are still other erroneous behavioral patterns that need to be calibrated.
Tool-Light pioneers the investigation into broader issues such as tool call insufficiency and reasoning redundancy~\cite{toollight}. Nevertheless, Tool-Light employs the DPO algorithm. This binary comparison optimization can cause the model output to collapse into a very narrow action space, resulting in limited exploration. It is worth noting that TIR scenarios feature an extremely extensive tool-use action space, which can give rise to a multitude of potential trajectories~\cite{se-agent,athoroughexamination,diversityofthought,optimizingpromptmontecarlotree,arpo}. Relying solely on imitation learning and binary comparison optimization, it is difficult to fully calibrate the agent's behavioral patterns. Therefore, how to utilize sufficient exploration within the TIR action space to effectively calibrate behavioral patterns via on-policy training remains an open challenge.

In this paper, we first conduct a comprehensive investigation into erroneous behavioral patterns within TIR tasks. By summarizing previous works and conducting quantitative analysis in preliminary experiments~\cite{traject-bench,butterflyeffectsinTIR,toolsfaildetecting,thereasoningtrap}, we categorize these error patterns into two primary aspects: \textbf{\textit{improper tool-use}} and \textbf{\textit{flawed reasoning logic}}, as shown in Figure~\ref{fig:intro}. Motivated by this analysis, we propose \ourmodel{}, an iterative, behavior-calibration training framework optimizing agent's TIR behavioral patterns at both the data and algorithmic levels. From data perspective, we introduce a \textbf{Self-evolving Data Flywheel}. We carefully design methods to augment existing trajectories. Training on these samples can broaden the tool-use action space coverage, enabling the agent to fully explore tool-use action space.

From an algorithmic perspective, we propose a \textbf{Behavior Calibration Training} framework with two phases. First, we implement Action Space Exploration Fine-tuning with the data from the flywheel, aiming to broaden the model's exploration of the tool-use action space. Based on this phase, we implement an Iterative Behavior Calibration Reinforcement Learning phase. This process alternates between Group-wise Pareto Sampling and Curriculum RL Training. The former identifies the most helpful data for training, while the latter gradually calibrates agent's actions toward optimal and standardized behavioral patterns. 

To verify the effectiveness of \ourmodel{}, we test multiple metrics on six challenging tasks. Overall, the results show that \ourmodel{} can achieve best performance in all dimensions, indicating the success of TIR behavior calibration.
In summary, our contributions are as follows:
\begin{itemize}[leftmargin=1em]
\item We provide a comprehensive quantitative analysis of erroneous behavioral patterns in TIR. Inspired by this, we propose \ourmodel{}, a framework for optimizing TIR's behavioral patterns.
\item We introduce a self-evolving data flywheel, an iterative loop where the model continuously refines its previous trajectories. This mechanism effectively unfolds the model's action space coverage beyond its initial exploration.
\item Based on the flywheel, we present a behavior calibration training framework with two phases, aiming to calibrate the model's exploration in tool-use action space to optimal trajectories.
\item Extensive experiments demonstrate that \ourmodel{} substantially improves behavioral efficiency, reasoning conciseness, and execution success rates while maintaining high accuracy.
\end{itemize}

\section{Related Work}
\label{sec:related work}
\paragraph{Tool-Integrated Reasoning.} Tool-Integrated Reasoning (TIR) empowers LLMs to solve complex tasks via external tools~\cite{stopoverthinking,CPT,towardsgeneralagentic,websailor,chainofcode,dparag}. Existing works generally endow LLM-based agents with TIR ability through: \textbf{(1) Inference-time Scaling}, focusing on prompt engineering or test-time computation scaling methods~\cite{search-o1,agenticreasoning,octotools,react}; and \textbf{(2) Training-based Method}, optimizing LLM policies using SFT, DPO, or RL methods~\cite{dotamath,start,agenticRLsurvey,tongyi,zhao2026roireasoningrationaloptimizationinference}. For example, works like Search-o1 guide agents to complete TIR tasks by designing workflows~\cite{search-o1}, while works like ToRL fully enhance agents' capabilities using RL methods~\cite{torl}. Currently, most works only focus on prioritizing accuracy, and only a few works pay attention to behavioral patterns' effectiveness such as the efficiency of TIR. In contrast, \ourmodel{} optimizes the behavioral patterns in TIR more comprehensively, using the on-policy training method to progressively calibrate TIR behavioral patterns.
\paragraph{Agentic Reinforcement Learning.}Recently, agentic reinforcement learning (RL) method plays an important role in TIR tasks~\cite{agenticRLsurvey,tongyi}. Early works focus on using RL methods to accomplish specific domain tasks~\cite{torl,search-r1,r1searcher,cort}. Building on this, some works make breakthroughs in more difficult TIR tasks or successfully expand the domain of TIR tasks~\cite{chainofcode,r1-code-intepreter,beyond10turns,chen2025iterresearchrethinkinglonghorizonagents}. For example, a series of works by Tongyi DeepResearch successfully improve agents' performance in deep search tasks~\cite{webdancer,webleaper,websailor,webshaper}. Works such as Tool-Star and SMART enhance the performance of TIR agents in tasks across a wider range of domains~\cite{dong2025toolstarempoweringllmbrainedmultitool,smart,autotir,ikea,vthinker,deepagent}. Recently, some RL algorithms or frameworks specifically adapted for agents have emerged~\cite{aepo,ggpo,treerl,treepo}. For instance, ARPO optimizes the branching strategy of GRPO from the perspective of information entropy, while VerlTool designs a training framework suitable for tool use~\cite{arpo,verltool}. Inspired by these works, \ourmodel{} successfully uses RL methods to optimize agents' tool-use behavioral patterns.


\begin{figure}[t]
  \includegraphics[width=\columnwidth]{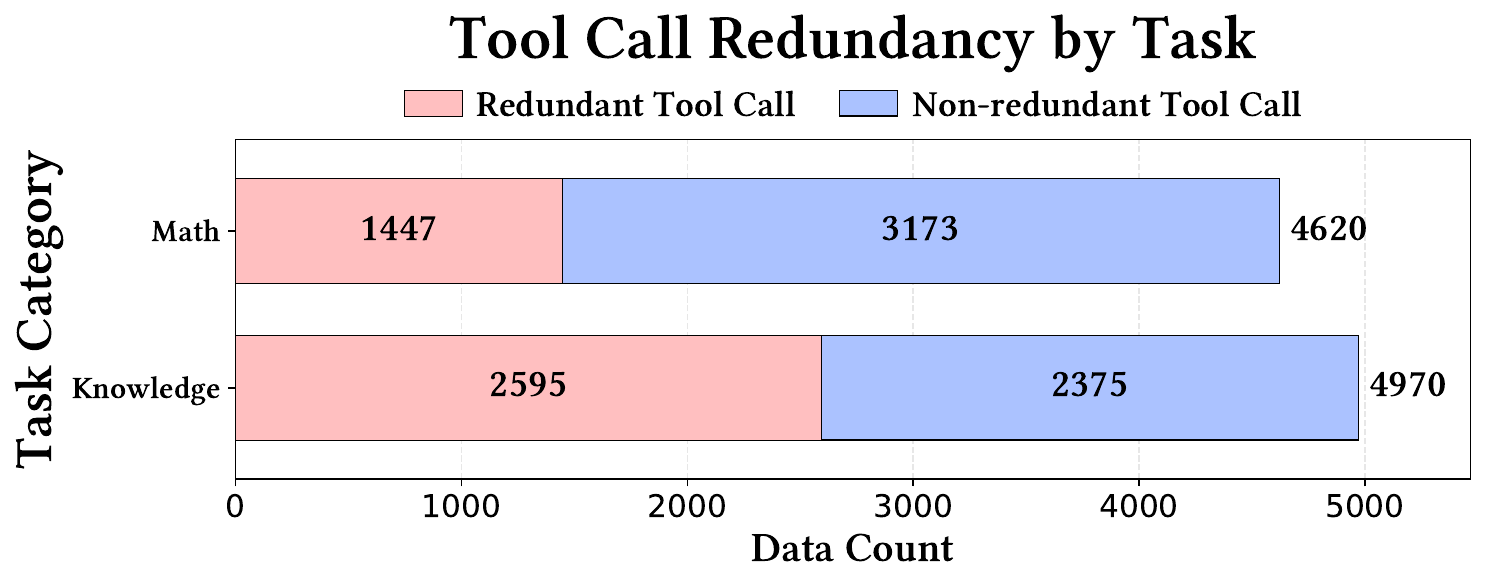}
  \caption{Statistics about redundant tool calls.}
  \label{fig:pre_redundant_tool_calls}
\end{figure}

\section{Preliminary Experiment}
\label{sec:Preliminary Experiment}

After summarizing the TIR behavioral patterns in previous works~\cite{traject-bench,butterflyeffectsinTIR,toolsfaildetecting}, we categorize erroneous behavioral patterns into two types:
\begin{itemize}[leftmargin=1em]
\item 
\textbf{Improper Tool-Use:} 
Incorrect tool triggering methods affect downstream tasks' performance. This includes \textbf{(1) Redundant Tool Calls}, where tool invocations fail to deliver additional useful information for solving the problem; and \textbf{(2) Aborted Tool Execution}, wherein tool execution is terminated due to improperly formulated tool calls generated by the agent--such as null query or defective code.
\item 
\textbf{Flawed Reasoning Logic}: 
Errors originating from defective cognitive processes. This includes \textbf{(1) Insufficient Tool Calls}, where the agent does not realize that additional tool calls are needed, leading to an incorrect result; and \textbf{(2) Erroneous Reasoning Process}, which involves logically irrelevant or erroneous steps derived from flawed planning. 
\end{itemize}





\begin{figure}[t]
  \includegraphics[width=\columnwidth]{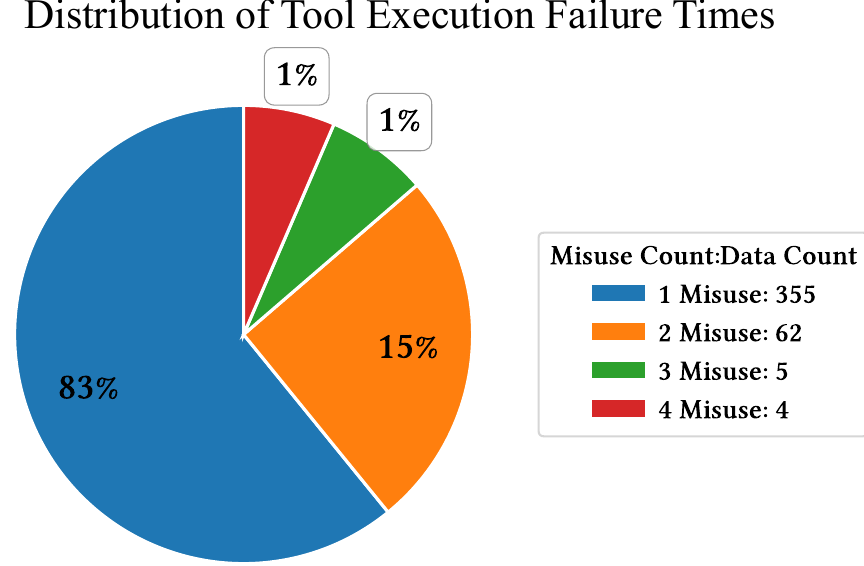}
  \caption{Statistics on Aborted Tool Execution Cases.}
  \label{fig:tool execution failure}
\end{figure}

\paragraph{Analysis of Improper Tool-Use.}
We first analyze redundant tool call situations. By employing GPT-4o~\cite{gpt4o} to identify redundant tool calls in outputs, as shown in Figure~\ref{fig:pre_redundant_tool_calls}, we observe a high prevalence of redundancy across both tasks, which highlights that uncalibrated agents struggle to call tools efficiently. We also analyze Aborted Tool Execution situations. As shown in Figure~\ref{fig:tool execution failure}, the majority of trajectories involve only one time error, while a small number of samples have errors occurring more than once. It suggests that improving triggering tool call accuracy through training is highly feasible.

\begin{figure}[t]
    \centering
    \begin{subfigure}[b]{0.51\linewidth}
        \centering
        \includegraphics[width=\linewidth]{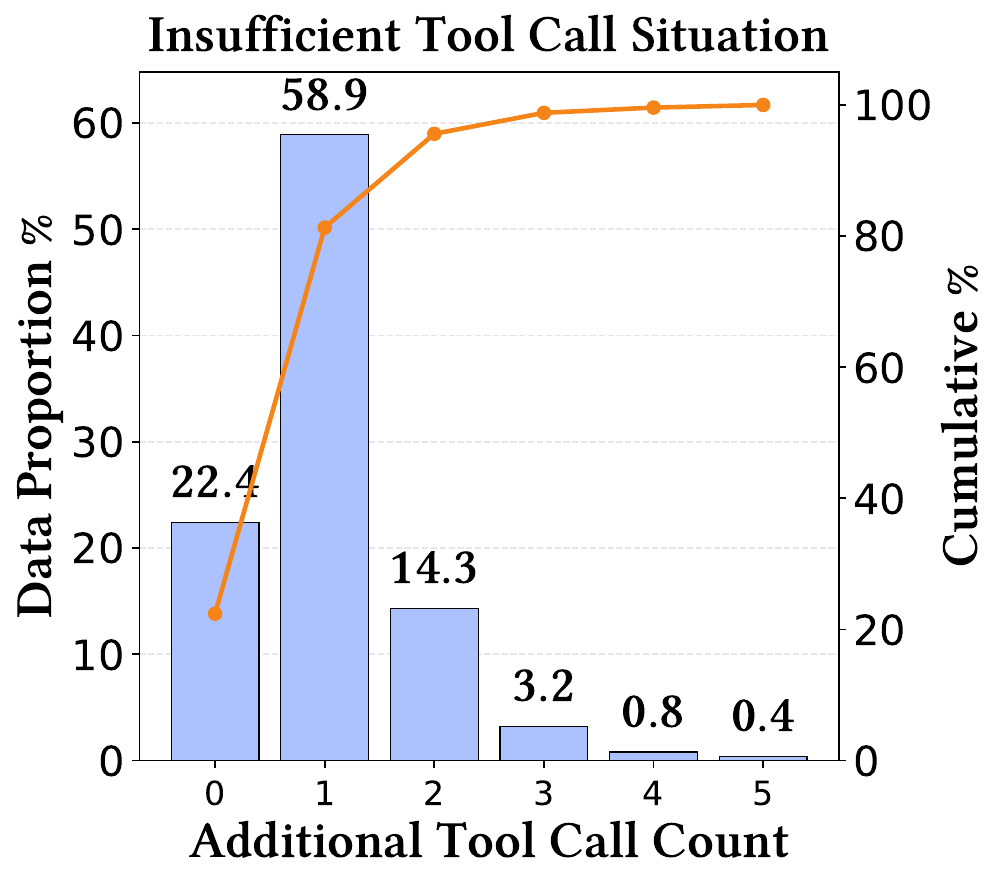}
        \caption{Additional tool calls needed for modification.}
        \label{fig:preliminary2_insufficient} 
    \end{subfigure}%
    \hfill
    \begin{subfigure}[b]{0.45\linewidth}
        \centering
        \includegraphics[width=\linewidth]{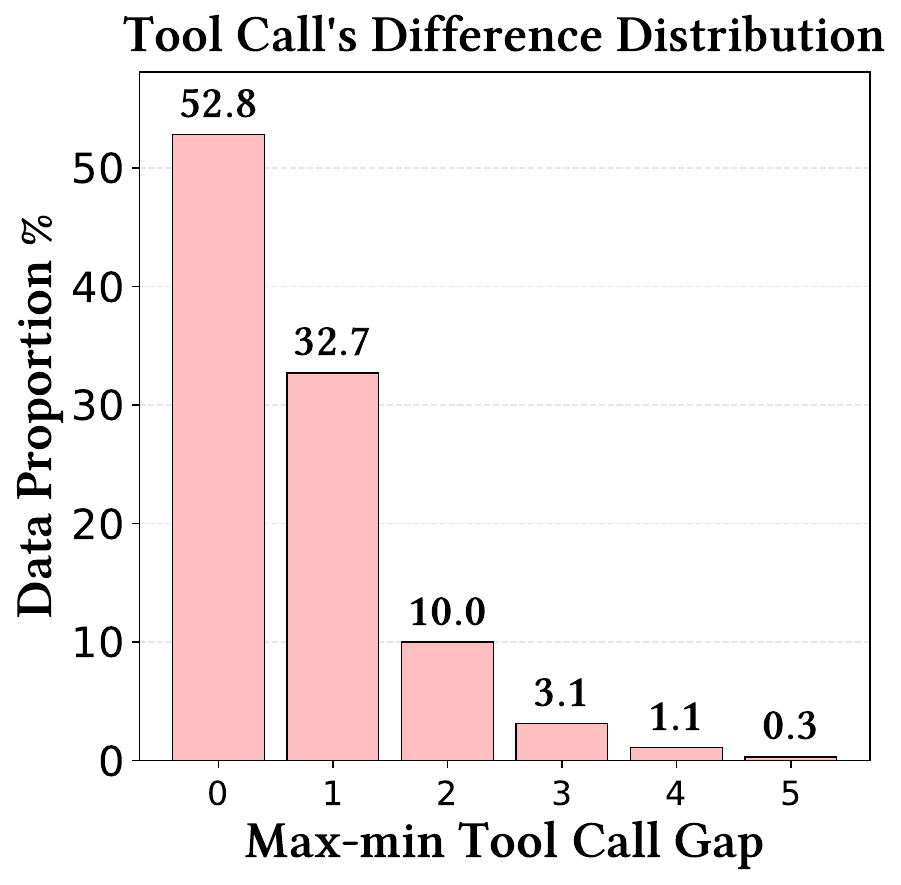}
        \caption{Distribution of different trajectorys' tool calls.}
        \label{fig:preliminary1_redundant} 
    \end{subfigure}
    \caption{Distribution of additional tool calls required to modify incorrect outputs, and differences in tool call across trajectories for identical questions.}
    \label{fig:preliminary12}
\end{figure}

\paragraph{Analysis of Flawed Reasoning Logic.}
To investigate insufficient tool call situations, we inject hints at the end of incorrect trajectories to induce continued reasoning. As shown in Figure~\ref{fig:preliminary12}(\subref{fig:preliminary2_insufficient}), a considerable number of trajectories can be modified after continued reasoning. It suggests that it is hard for the agent to invoke additional tools, resulting in the ``premature closure'' issue.

\paragraph{Analysis of Trajectories' Diversity.}
Finally, we investigate the intrinsic complexity of the TIR action space by analyzing the diversity among correct solutions. Figure~\ref{fig:preliminary12}(\subref{fig:preliminary1_redundant}) reveals substantial disparities in tool call frequency even across correct outputs. This observation confirms that the action space for TIR tasks is extensive. Agents without calibration can hardly explore an effective reasoning trajectory.

Full dataset details and error case studies are in Appendix~\ref{app:dataset} and~\ref{app:case study}, and full experiment details can be found in Appendix~\ref{app:pre exp details}. 
Examples for each type of error can be found in Appendix~\ref{app:case study}.

\section{The Proposed Method: \ourmodel{}}

\begin{figure*}
    \centering
    \includegraphics[width=\linewidth]{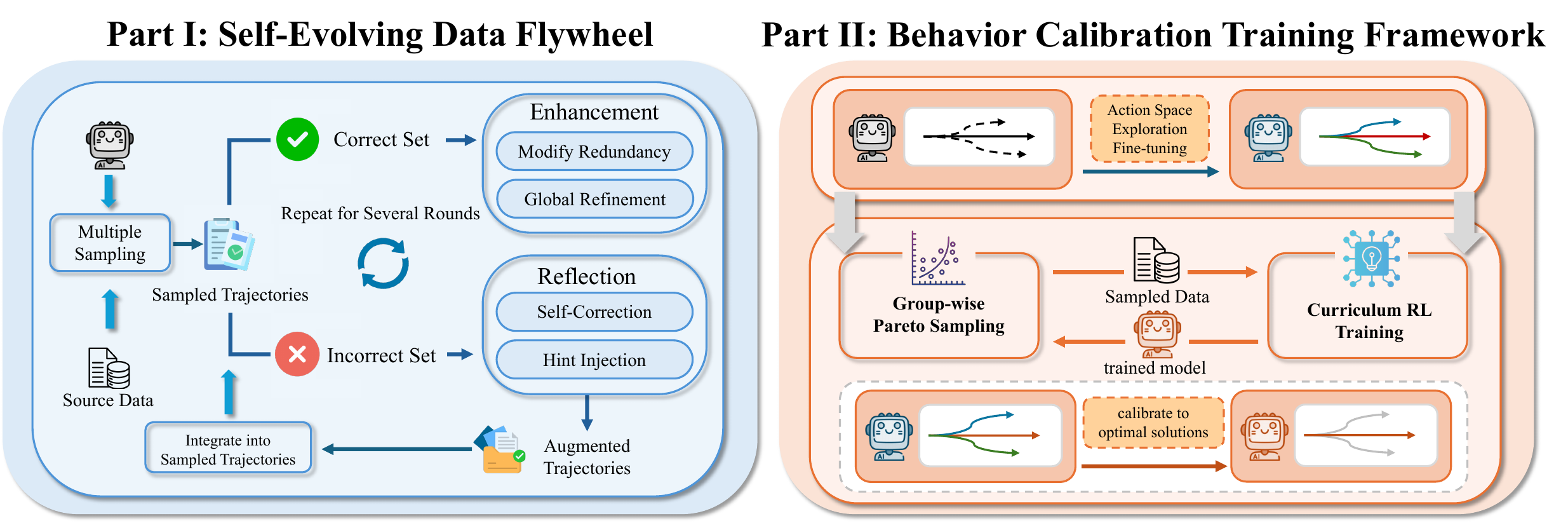}
    \caption{The overall process of \ourmodel{}. The left part represents Self-Evolving Data Flywheel, while the right part represents Behavior Calibration Training Framework.}
    \label{fig:framework}
\end{figure*}

We introduce the \ourmodel{} framework, aiming to calibrate LLM-based agent's TIR behavioral correctness. As shown in Figure~\ref{fig:framework}, \ourmodel{} comprises two primary aspects: 
\textbf{(1) Self-Evolving Data Flywheel} 
obtains rich and diverse training data through self-corrective refinement or expanded exploration to iteratively optimize trajectories.
\textbf{(2) Behavior Calibration Training} conducts a two-phase training framework to calibrate TIR behavioral patterns.

\subsection{Self-Evolving Data Flywheel}


In order to more effectively calibrate the agent's behavior to the optimal patterns, it is necessary for the agent to fully explore the action space.
Inspired by LLM Self-Evolution and Test-Time Scaling methods~\cite{asurveyofselfevolvingagents,autoif,selfchallenginglanguagemodelagents,qiao2025we}, we propose a Self-evolving Data Flywheel to enrich training dataset. Specifically, the agent is iteratively guided to either refine generated trajectories or explore diverse solution paths, merging valid samples back into the pool. This mechanism can yield diverse, high-quality trajectories for fine-tuning. 

\paragraph{Initialization.} For each question $q$ in the original data $D_{\text{source}}$, we use prompt $I_1$ and guide LLM (defined as $M$) in generating multiple outputs, establishing an initial action distribution. Based on answer correctness, we divide these outputs (defined as $D_{\text{rollout}}^q$) into \textit{Correct Set} and \textit{Incorrect Set}, serving as the foundation for the flywheel.
\paragraph{Correct Reasoning Enhancement.} For the \textit{Correct Set}, we design two enhancement strategies to remove redundant thinking steps or tool calls.
\textbf{(1) Redundant Modification:} According to recent studies~\cite{lostatthebegining, toollight}, we specifically refine the first redundant tool call. We define a ``step'' as the content generated by the LLM between the previous tool execution result (or initial query) and the subsequent tool call (or final answer). We use prompt $I_2$ to instruct $M$ to identify the first step with redundancy, generate a refined step, and regenerate the subsequent path from it.
\textbf{(2) Global Refinement:} Complementarily, we design prompt $I_3$ to instruct $M$ to refine all thinking processes while preserving the logical consistency of the tool call sequence.

\paragraph{Incorrect Reasoning Reflection.} For the \textit{Incorrect Set}, we also design two enhancement strategies for modification.
\textbf{(1) Self-Correction:} We design prompt $I_4$ and instruct $M$ to find the first incorrect reasoning step, modify it, and continue reasoning.
\textbf{(2) Hint Injection:} As described in Section~\ref{sec:Preliminary Experiment}, agents without behavior calibration may occur insufficient tool call issues. START guides the agent in reasoning with tool calls by inserting hints into the reasoning chain~\cite{start}. Inspired by this, we design different hints for different tasks, and insert them after the identified error step or at the end of the whole trajectory. This operation aims to encourage $M$ to call additional tools to assist in reasoning and derive alternative solutions.

\paragraph{Iterative Data Evolution.} We reintegrate enhanced trajectories into $D_{\text{rollout}}^q$ and repeat this process for $R$ iterations, yielding the final dataset $D_{\text{aug}}$. It significantly broadens the action space coverage of initial dataset. It can guide $M$ to learn comprehensive exploration strategies during the fine-tuning phase, which establishes a robust foundation for subsequent group-wise behavior calibration. The algorithm is outlined in Algorithm~\ref{alg:self-evolve-flywheel}, and prompts and hints can be found in Appendix~\ref{app:prompt}.

\begin{algorithm}[t]
\small
\caption{Self-Evolving Data Flywheel}
\textbf{Input} initial policy model $M$; original training data $D_{\text{source}}$; Maximum evolution rounds $R$
\begin{algorithmic}[1]
\State Initialize $D_{\text{aug}}=\emptyset$
\For{$q$ in $D_{\text{source}}$} 
    \State $D_{\text{rollout}}^q=M (I_1, q)$
    \For{$r$ = 1, \dots, $R$}
        \State Divide $D_{\text{rollout}}^q$ into Correct Set, Incorrect Set
        \State $\text{Trajs}^1=\text{Enhancement}(\text{Correct Set})$
        \State $\text{Trajs}^2=\text{Reflection}(\text{Incorrect Set})$
        \State $D_{\text{rollout}}^q$.\text{append}($\text{Trajs}^1, \text{Trajs}^2$)
    \EndFor 
    \State $D_{\text{aug}}.\text{append}(D_{\text{rollout}}^q)$
\EndFor 
\end{algorithmic}
\textbf{Output} $D_{\text{aug}}$
\label{alg:self-evolve-flywheel}
\end{algorithm}




\subsection{Behavior Calibration Training Framework}

In this section, we introduce a Behavior Calibration Training framework. First, we conduct an \textbf{Action Space Exploration Fine-tuning}. Then, we conduct \textbf{Iterative Behavior Calibration Reinforcement Learning}. It operates by alternating between Group-wise Pareto sampling and Curriculum RL Training. Through this training framework, the model's tool call actions in the action space can be calibrated to optimal behavior patterns.

\subsubsection{Action Space Exploration Fine-tuning}
We first impose quality controls on $D_{\text{aug}}$, discarding trajectories with incorrect answers, formatting errors, or execution failures. Then we perform rejection sampling fine-tuning (RFT) on $M$~\cite{rft}. The training target can be expressed as: 
$\max_{\theta} \mathbb{E}_{(x, y) \sim \mathcal{D}} \left[ \sum_{t=1}^{|y|} \log P_\theta(y_t \mid x, y_{<t}) \right].$
RFT on flywheel-enhanced trajectories expands the agent's exploration of the action space, while quality controls calibrate the agent's behavior to eliminate Aborted Tool Execution and format errors. Driven by expanded exploration, trained $M$ exhibits pronounced group-wise trajectory diversity, providing the necessary variance for effective group-wise calibration in the subsequent RL phase.

\subsubsection{Group-wise Pareto Sampling}

Traditional group-based RL methods are susceptible to trajectory homogenization, leading to vanishing gradients~\cite{dagrpo,deepseek-r1,aepo}. To overcome this, drawing on existing works~\cite{nsga-ii,geneticalgorithms}, we introduce a Pareto Sampling strategy. By strictly selecting samples from the Pareto frontier which preserves enough diversity, this approach ensures significant reward differentials within groups, preventing gradient decay and calibrating the agent toward high-quality trajectories.

\paragraph{Group-wise Dispersion Definition}
To quantify the distributional divergence, we sample $K$ trajectories $\{\tau_i\}_{i=1}^K$ for $q$ in $D_{\text{source}}$ (with $K=16$ to approximate the real distribution). To effectively calibrate the agent's behavioral patterns while further improving outcome correctness, we construct two complementary sampling indices:
\textbf{(1) Correctness Dispersion ($S_{\text{corr}}$)} is defined as the standard deviation of trajectory correctness scores: $S_{\text{corr}}(x) = \sqrt{\frac{1}{K} \sum_{i=1}^{K} (r(\tau_i) - \bar{r})^2}$, where $\bar{r}$ is the mean score. A higher $S_{\text{corr}}$ implies greater potential for optimization in problem-solving accuracy.
\textbf{(2) Behavioral Dispersion ($S_{\text{tool}}$)} is defined as the standard deviation of tool invocation counts: $S_{\text{tool}}(x) = \sqrt{\frac{1}{K} \sum_{i=1}^{K} (l(\tau_i) - \bar{l})^2}$, where $\bar{l}$ is the mean invocation count. High $S_{\text{tool}}$ implies rich TIR signals for behavioral exploration.
\paragraph{Sample Selection Strategy}
We define the dominance relationship between samples. A sample $p$ dominates $q$ ($p \succ q$) if $p$ is non-inferior in all indices and superior in at least one. 
Based on this, we implement a two-stage selection strategy.
\textbf{(1) Fast Non-dominated Sorting:} We partition the dataset $D_{\text{source}}$ into Pareto frontiers $\mathcal{F}_1, \mathcal{F}_2, \dots$, where $\mathcal{F}_1$ contains non-dominated optimal solutions, and $\mathcal{F}_{i+1}$ are the suboptimal solutions only dominated by $\mathcal{F}_i$. We prioritize selecting samples from higher frontiers to maximize gradient contribution.
\textbf{(2) Crowding Distance Truncation:} When adding frontier $\mathcal{F}_i$ exceeds the target size $N$, we calculate the crowding distance in this frontier. We sort samples along $S_{\text{corr}}$ and $S_{\text{tool}}$ dimensions, and compute crowding distances based on adjacent intervals. We prioritize retaining samples with larger crowding distances to maximize diversity. \footnote{The specific sampling flow can be found in Appendix~\ref{app:pareto sampling}.}

\subsubsection{Curriculum RL Training}
In this section, we present a Curriculum RL framework to calibrate TIR behaviors. We employ group-wise policy optimization with a reward mechanism. This can effectively calibrates behavior patterns while strictly preserving reasoning correctness.
\paragraph{Reward Mechanism Design.}
To maintain format, outcome, and thinking process accuracy while calibrating behavioral patterns, we employ a multi-objective group-wise reward mechanism. Specifically, we design four key components to target these distinct objectives: format reward, correctness reward, TIR behavior pattern reward, and thinking process reward.

\begin{itemize}[leftmargin=1em]
\item \textbf{Format and Correctness Rewards.} Following prior works~\cite{dong2025toolstarempoweringllmbrainedmultitool,torl,search-r1}, we define the correctness reward ($R_{\text{corr}}$), employing the F1 score for knowledge-intensive task, and a binary reward for mathematical reasoning task.
We also impose a formatting constraint ($R_{\text{format}}$), with -$1$ reward for format-incorrect outputs.
\item \textbf{Efficiency-driven Behavior Rewards.} 
We introduce rewards for tool call and reasoning length to penalize inefficient behaviors. For trajectory $i$, the tool call score $f_k$ is as follows:
\begin{equation}
\small
f_{\text{tool}} = \frac{1}{1 + e^{\sigma_{\text{tool}} \cdot (v_{i,\text{tool}} - \bar{v}_\text{tool})}}.
\label{eq:tool-reward}
\end{equation}

The reasoning length score is as follows:
\begin{equation}
\small
f_{\text{len}} = \frac{1}{1 + e^{\sigma_{\text{len}} \cdot (v_{i,\text{len}} - \bar{v}_\text{len})}}.
\label{eq:len-reward}
\end{equation}

Among them, $v_{i,\text{tool}}$, $\bar{v}_\text{tool}$, $\sigma_{\text{tool}}$, $v_{i,\text{len}}$, $\bar{v}_\text{len}$, and $\sigma_{\text{len}}$ represent the tool call count for trajectory $i$, the group average tool call count, the tool call hyperparameter, the reasoning length, the group average reasoning length, and the reasoning length hyperparameter, respectively.
For group-wise reasoning lengths, they do not include tool execution results, and we perform Z-score standardization among them. This mechanism incentivizes efficiency: by assigning higher rewards to samples with superior indicators, it explicitly steers the agent to more concise reasoning paths.

\begin{table*}[t]
\small
\centering
\caption{
Results on six tasks, with top two results highlighted in \textbf{bold} and \underline{underlined}. 
\myicon{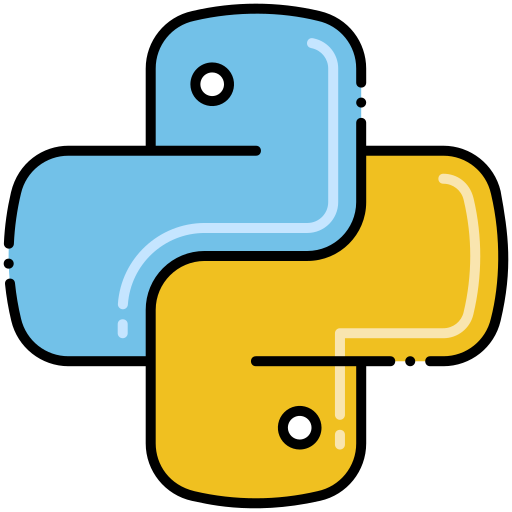} and \myicon{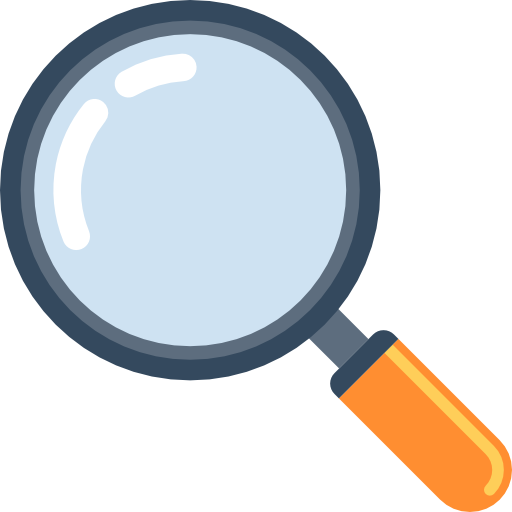} represent method with code compiler and search tool, respectively.
For fairness, we retrain a WebSailor-7B model using Wikipedia-based data. Abbreviation: it (Instruct), 2Wiki (2WikiMultiHopQA), Bamb (Bamboogle), MSQ (MuSiQue).}
\label{tab:main}
\setlength{\tabcolsep}{2mm} 
\begin{tabular}{lcccccccccccccc}
\toprule
 \multirow{3}{*}[-1.5ex]{\textbf{Method}} & \multicolumn{6}{c}{\textbf{Mathematical-Reasoning Tasks}} & \multicolumn{6}{c}{\textbf{Knowledge-Intensive Tasks}} & \multicolumn{2}{c}{\multirow{2}{*}{\textbf{Avg.}}} \\
\cmidrule(lr){2-7}\cmidrule(lr){8-13}
& \multicolumn{2}{c}{\textbf{AIME24}} & \multicolumn{2}{c}{\textbf{AMC23}} & \multicolumn{2}{c}{\textbf{MATH500}} & \multicolumn{2}{c}{\textbf{2Wiki}} & \multicolumn{2}{c}{\textbf{Bamb}} & \multicolumn{2}{c}{\textbf{MSQ}} & \\ 
\cmidrule(lr){2-3}\cmidrule(lr){4-5} \cmidrule(lr){6-7} \cmidrule(lr){8-9} \cmidrule(lr){10-11} \cmidrule(lr){12-13} & LJ & Effi & LJ & Effi & LJ & Effi & F1 & Effi & F1 & Effi & F1 & Effi & P & Effi \\
\midrule
\multicolumn{12}{l}{\textit{\textbf{Direct Inference}}} \\
Qwen2.5-7B-it & 10.0 & - & 30.0 & - & 57.2 & - & 18.9 & - & 24.5 & - & 8.5 & - & 24.9 & - \\
\midrule
\multicolumn{12}{l}{\textit{\textbf{Single-TIR Methods}}} \\
Search-o1 \myicon{latex/figures/search.png} & 10.0 & 10.0 & 37.5 & 37.5 & 65.4 & 65.2 & 42.9 & 26.2 & 48.5 & 33.9 & 13.3 & 10.9 & 36.3 & 30.6\\
Search-R1 \myicon{latex/figures/search.png} & 23.3 & 18.3 & 35.0 & 25.6 & 57.8 & 39.3 & 49.2 & 15.5 & 40.5 & 13.7 & 18.6 & 4.8 & 37.4 & 19.6\\
Research \myicon{latex/figures/search.png} & 10.0 & 5.8 & 52.5 & 21.7 & 64.0 & 31.2 & \underline{67.1} & 25.5 & 58.4 & 24.8 & 27.8 & 9.3 & 46.6 & 19.7 \\
WebThinker \myicon{latex/figures/search.png} & 16.7 & 16.7 & 45.0 & 45.0 & 66.0 & 65.7 & 60.8 & 24.8 & 46.3 & 23.4 & 18.6 & 9.1 & 42.2 & 30.8 \\
WebSailor* \myicon{latex/figures/search.png} & 26.7 & 2.9 & 65.0 & 23.5 & 72.2 & 13.9 & 61.5 & 5.3 & 58.4 & 6.1 & 17.0 & 1.6 & 50.1 & 8.9\\
ToRL \myicon{latex/figures/python.png} & \underline{40.0} & \underline{40.0} & \underline{75.0} & \underline{71.1} & \textbf{84.6} & \textbf{83.8} & 35.1 & \underline{35.1} & 14.2 & 14.2 & 5.8 & 5.8 & 42.5 & \underline{41.6} \\
DotaMath \myicon{latex/figures/python.png} & 16.7 & 16.7 & 50.0 & 50.0 & 74.6 & 72.5 & 16.3 & 16.3 & 18.9 & 18.9 & 7.6 & 7.6 & 30.7 & 30.3 \\
START \myicon{latex/figures/python.png} & 23.3 & 23.3 & 67.5 & 62.1 & 78.4 & 74.8 & 15.6 & 15.6 & 19.8 & 19.8 & 5.7 & 5.7 & 35.1 & 33.6 \\
\midrule
\multicolumn{12}{l}{\textit{\textbf{Multi-TIR Methods}}} \\
Prompting & 13.3 & 6.1 & 47.5 & 33.54 & 67.0 & 51.7 & 18.0 & 10.5 & 17.9 & 12.4 & 8.5 & 5.6 & 28.7 & 20.0\\
IKEA & 10.0 & 6.1 & 42.5 & 32.0 & 63.6 & 44.2 & 34.7 & 9.7 & 35.8 & 11.2 & 14.8 & 3.8 & 33.6 & 17.8\\
SMART & 3.3 & 3.3 & 17.5 & 15.8 & 39.4 & 33.0 & 21.7 & 21.1 & 40.7 & \textbf{40.6} & 12.3 & 11.8 & 22.5 & 20.9\\
AutoTIR & 33.3 & 21.3 & 62.5 & 46.4 & 72.6 & 61.2 & 66.2 & 27.8 & 56.9 & 29.5 & \textbf{30.9} & 12.0 & 53.7 & 33.1\\
Tool-Star & 33.3 & 25.6 & 60.0 & 53.8 & 80.4 & 70.0 & 65.8 & 29.0 & 55.4 & 34.2 & 24.5 & \underline{12.6} & 52.3 & 37.5\\
Tool-Light & 33.3 & 33.3 & 72.5 & 67.5 & 79.0 & 72.8 & 60.4 & 24.6 & \underline{58.7} & 23.0 & 26.0 & 9.8 & \underline{55.0} & 38.5\\
\midrule
\ourmodel{} (Ours) & \textbf{46.7} & \textbf{43.3} & \textbf{77.5} & \textbf{71.3} & \underline{81.6} & \underline{76.1} & \textbf{67.2} & \textbf{35.6} & \textbf{59.4} & \underline{34.4} & \underline{28.0} & \textbf{15.3} & \textbf{60.1} & \textbf{46.0}\\
\bottomrule
\end{tabular}
\end{table*}

\end{itemize}

Overall, the reward for trajectory $i$ aggregates these components:
\begin{equation}
\small
R_i = R_{\text{corr}}^{i} \cdot f_{\text{tool}}^{i} \cdot f_{\text{len}}^{i} + R_{\text{format}}^{i}.
\label{eq:reward}
\end{equation}
By maximizing rewards for correct, efficient, and well-formatted samples, this mechanism compels calibration towards superior trajectories, promoting overall behavioral effectiveness during training.

\paragraph{ARPO Training.}
We employ Agentic Reinforced Policy Optimization (ARPO) for RL~\cite{arpo}. The objective function is:
\begin{equation}
\footnotesize
\begin{split}
    \mathcal{J}_{\text{ARPO}}(\theta) = \mathbb{E}_{\left[ q \sim D, \{ o_i \}_{i=1}^G \sim \pi_{\theta_{\text{old}}}(\cdot | q) \right]} \\ 
    & \hspace*{-15.5em} \left[ \frac{1}{G} \sum_{i=1}^G \frac{1}{|o_i|} \sum_{t=1}^{|o_i|} \min \left( r_t(\theta) A_{i}, \text{clip} \left( r_t(\theta), 1 \pm \epsilon \right) A_{i} \right) \right],
\end{split}
\label{eq:grpo}
\end{equation}
where $r_t(\theta)$ is the probability ratio, and $\epsilon$ is the clipping hyperparameter. The advantage $A_{i}$ for the $i$-th trajectory is derived from the normalized rewards within the group:
$A_{i} = \frac{R_i - \bar{\textbf{R}}}{\text{std}(\textbf{R})}$.
Drawing on curriculum learning strategies~\cite{toollight,webthinker,r1-code-intepreter}, we alternate between Pareto sampling and ARPO training over multiple rounds. To prevent reward hacking, we progressively decrease $\sigma_{\text{tool}}$ and $\sigma_{\text{len}}$ in each round. This iterative process gradually guides the model's exploration toward optimal trajectories, stabilizing both behavioral patterns and answer accuracy.

\section{Experiments}
\subsection{Experimental Setup}
\paragraph{Tool Design. }According to existing works~\cite{search-o1,torl,dong2025toolstarempoweringllmbrainedmultitool,toollight}, we design two tools to assist in reasoning tasks: \textbf{Search Engines} (including local search and web search) retrieve and summarize external knowledge, and \textbf{Code Compiler} executes LLM-generated code. These tools are sufficient for supporting tasks evaluated in this study.
\paragraph{Datasets} 
We conduct evaluations on two types of 
reasoning tasks: (1) Mathematical-Reasoning tasks (including AIME24,\footnote{https://huggingface.co/datasets/AI-MO/aimo-validation-aime} AMC23,\footnote{https://huggingface.co/datasets/zwhe99/amc23} MATH500~\cite{math500}) and (2) Knowledge-Intensive tasks (including 2WikiMultiHopQA~\cite{2wiki}, Bamboogle~\cite{bamboogle}, and MuSiQue~\cite{musique}).

\paragraph{Evaluation Metrics} We measure \ourmodel{}'s performance on correctness and efficiency aspects. For correctness, we utilize LLM-as-Judge~\cite{llm-as-judge} (Qwen2.5-72B-Instruct~\cite{qwen2.5} for Mathematical-Reasoning tasks) and F1 score (for Knowledge-Intensive tasks). Efficiency metric is calculated as the average correctness per tool call: $Effi = \frac{1}{N} \sum_{i=1}^{N} \frac{P_i}{T_i}$, where $P_i$ is the correctness score and $T_i$ is the tool call count for sample $i$.

\paragraph{Baselines} We categorize baselines into three groups: \textbf{(1) Direct Reasoning}, utilizing Qwen2.5-7B-Instruct~\cite{qwen2.5} without tools; \textbf{(2) Single-TIR methods}, including Search-o1, Search-R1, Research, WebThinker, WebSailor, ToRL, DotaMath, and START~\cite{search-o1,search-r1,webthinker,torl,start,research,websailor,dotamath}; and \textbf{(3) Multi-TIR methods}, including prompt-based approaches, IKEA, SMART, AutoTIR, Tool-Star, and Tool-Light~\cite{ikea,smart,autotir,dong2025toolstarempoweringllmbrainedmultitool,toollight}.

\paragraph{Implementation Details} Our training is all based on the Wikipedia dataset. During testing, for Knowledge-Intensive tasks, we use local retrieval based on Wikipedia documents. For Mathematical-Reasoning Tasks, we use Google Search. More details can be found in Appendix~\ref{sec:appendix-dataset-baseline} and~\ref{app:implementation details}.

\subsection{Main Results}
Main results are shown in Table~\ref{tab:main}. In correctness and efficiency aspects, \ourmodel{} achieves the best performance in most tasks. Through further analysis, we can draw the following conclusions:

\begin{figure}[t]
  \includegraphics[width=\columnwidth]{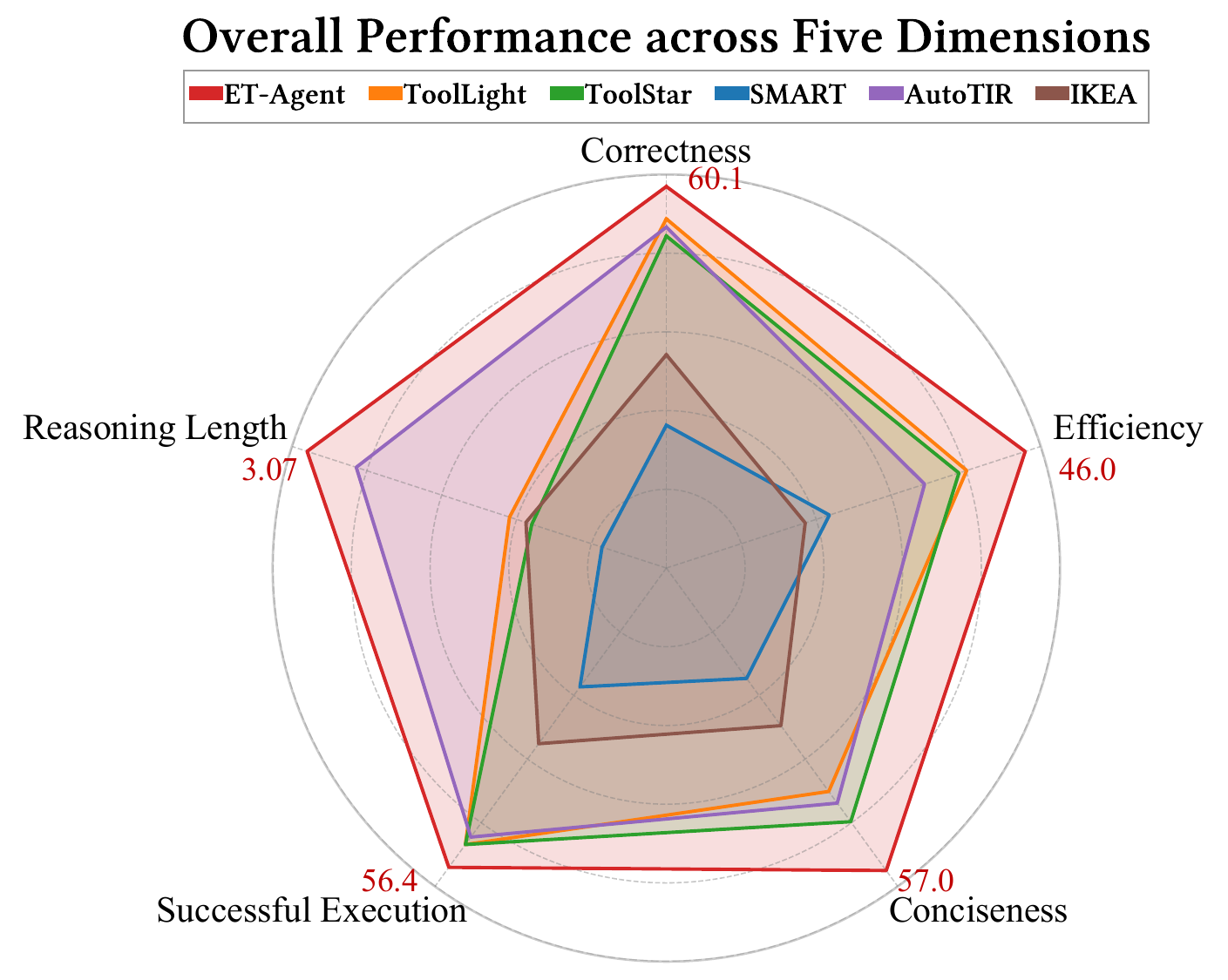}
  \caption{Performances across five metrics.}
  \label{fig:radar}
\end{figure}


\textbf{The Necessity of Training.} Models without training struggle to arrive at correct answers. For instance, the average correctness of Search-o1 is lower than that of Research by 10.3 points. These findings underscore the importance of training.

\textbf{Efficiency Neglection in Current Baselines.} Existing methods often struggle to balance accuracy with efficiency. For example, although Tool-Star achieves high correctness (52.3), its efficiency is surpassed by ToRL and Tool-Light. Similarly, while AutoTIR attains the highest correctness on MSQ, its efficiency (12.0) is significantly inferior to that of \ourmodel{} (15.3). This discrepancy highlights that many current TIR approaches neglect the optimization of behavioral patterns.

\textbf{Superior Performance of \ourmodel{}.} On average, \ourmodel{} achieves the best results in both correctness and efficiency indices (60.1, 46.0). It significantly outperforms baseline methods on datasets such as AIME24, AMC23, and 2Wiki, while maintaining a leading position across others. These results validate that our \ourmodel{} pipeline effectively enhances reasoning accuracy while standardizing the model's behavioral patterns.
\subsection{Tool-use Behavioral Patterns Analysis}
\label{sec:indepth analyze}
To comprehensively analyze the effectiveness of \ourmodel{}, we additionally define three metrics:
\textbf{(1) Conciseness} measures the brevity of tool calls. It is defined as $N_{\text{nr}} / N$, where $N_{\text{nr}}$ denotes the number of samples without redundant tool calls.
\textbf{(2) Successful Execution} evaluates the reliability of tool calls. It is defined as $\frac{1}{N} \sum_{i=1}^{N} \frac{P_i}{\text{Wrong}_i}$, where $\text{Wrong}_i$ represents the number of incorrect tool executions in sample $i$.
\textbf{(3) Reasoning Length} quantifies the compactness of the thought process. It is calculated as $\frac{1}{N} \sum_{i=1}^{N} \frac{P_i}{L_i}$, where $L_i$ is the token length of the reasoning chain (excluding tool execution results).

\begin{figure}[t]
  \includegraphics[width=\columnwidth]{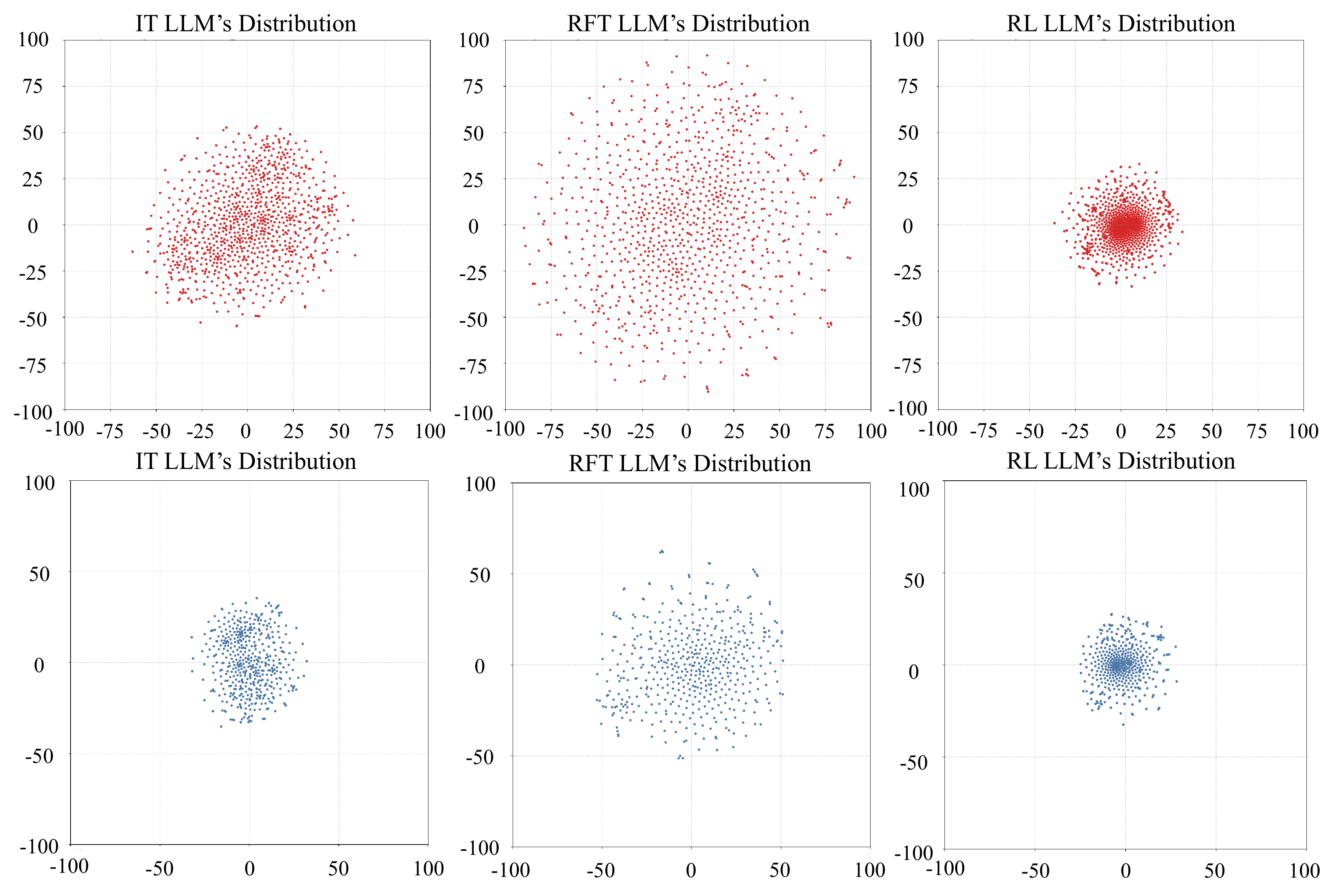}
  \caption{Distribution of action space exploration in different training stages. The more dispersed the distribution, the higher the diversity. \textbf{Top:} Qwen2.5-7B-it. \textbf{Bottom:} Llama3.1-8B-it.}
  \label{fig:tsne}
\end{figure}

\begin{figure}[t]
  \includegraphics[width=\columnwidth]{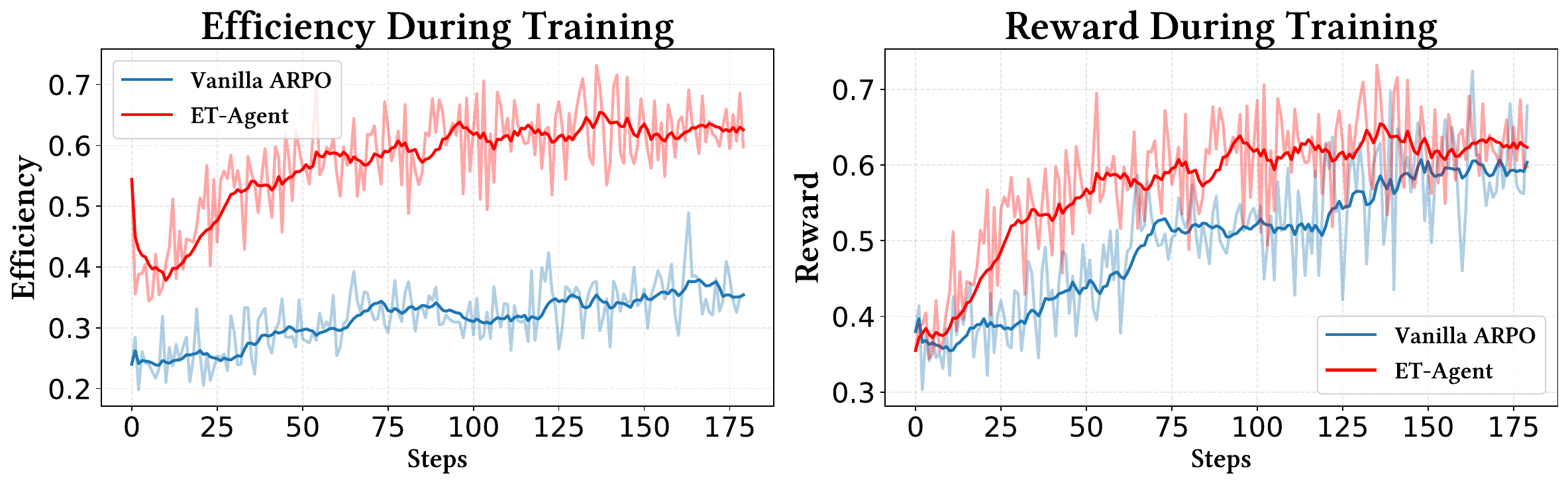}
  \caption{Comparison of efficiency and reward between vanilla ARPO and \ourmodel{} during the RL phase (only the final RL phase of \ourmodel{} is chosen).}
  \label{fig:training}
\end{figure}

Figure~\ref{fig:radar} illustrates the performance across these dimensions. \ourmodel{} achieves state-of-the-art results on all five metrics, significantly outperforming mainstream Multi-TIR baselines. This comprehensive superiority validates that our pipeline effectively standardizes the agent's behavioral patterns while ensuring robust reasoning.\footnote{Full evaluation results can be found in Appendix~\ref{app:full results of radar}.}

\subsection{Quantitative Analysis}
\paragraph{Action Space Distribution Analysis.}
We analyze the agent's output distributions across different training stages. We sample ten trajectories per query across three stages: the base Instruct LLM, the RFT LLM, and the RL LLM. All outputs are encoded using the e5 model~\cite{e5} and projected for visualization via t-SNE~\cite{t-sne}, as shown in Figure~\ref{fig:tsne}. The transition from the IT to RFT exhibits a more dispersed distribution, signifying expanded exploration of the action space. Subsequently, the RL phase yields a more compact distribution, indicating that the LLM converges this broad exploration into precise, optimal trajectories.
\paragraph{Training Dynamic Analysis.}We analyze the training dynamics by tracking efficiency and reward changes. We compare \ourmodel{} with a `Vanilla ARPO' baseline, which runs ARPO solely for result correctness. As shown in Figure~\ref{fig:training}, \ourmodel{} outperforms that baseline in both efficiency and reward metrics, thereby validating the effectiveness of our behavior pattern reward design.

\subsection{Ablation Study}

To fully explore the role of each component within the \ourmodel{} framework, we conduct ablation experiments, specifically focusing on the Self-evolving Data Flywheel, Group-wise Pareto sampling, and the reward mechanism design. As shown in Table~\ref{tab:ablation}: \textbf{(1)} Eliminating any component leads to performance degradation across both correctness and efficiency, validating the rationality of \ourmodel{} design. \textbf{(2)} Data quality is critical, as both the Flywheel and Pareto Sampling significantly outperform their respective baselines. \textbf{(3)} The reward design is essential; removing the group-wise reward mechanism lowers efficiency, while fixing $\sigma$ causes severe correctness degradation (indicative of reward hacking). This confirms that all components in our reward mechanism are indispensable for robust optimization.




\begin{table}[t]
\small
\centering
\caption{
Results of ablation experiments on different components, with top two results highlighted in \textbf{bold} and \underline{underlined}. Abbreviation: MR (Mathematical-Reasoning Tasks), KI (Knowledge-Intensive Tasks).}
\label{tab:ablation}
\setlength{\tabcolsep}{1.5mm} 
\begin{tabular}{lcccccc}
\toprule
 \multirow{2}{*}[-1.5ex]{\textbf{Method}} & \multicolumn{2}{c}{\textbf{MR}} & \multicolumn{2}{c}{\textbf{KI}} & \multicolumn{2}{c}{\textbf{Avg.}} \\
\cmidrule(lr){2-3}\cmidrule(lr){4-5}\cmidrule(lr){6-7} & LJ & Effi & LJ & Effi & LJ & Effi \\
\midrule
Qwen2.5-7B-it & 42.6 & 30.4 & 14.8 & 9.5 & 28.7 & 20.0\\
SFT & \textbf{58.0} & - & \textbf{40.6} & - & \textbf{49.3} & -\\
\quad {w/o Flywheel} & \underline{54.8} & - & \underline{39.7} & - & \underline{47.3} & -\\
\midrule
\ourmodel{} & \textbf{68.6} & \textbf{63.6} & \textbf{51.5} & \underline{28.4} & \textbf{60.1} & \textbf{46.0}\\
\quad {w/o Pareto} & 59.3 & 54.4 & 48.8 & 25.8 & 54.1 & 40.1\\
\quad w/o {Reward} & \underline{66.6} & 59.4 & \underline{50.5} & 27.3 & \underline{58.5} & \underline{43.4}\\
\quad w/o {$\sigma$ Decrease} & 62.2 &\underline{59.5} & 33.9 & \textbf{32.4} & 48.1 & \textbf{46.0}\\
\bottomrule
\end{tabular}
\end{table}

\section{Conclusion}

In this paper, we introduce \ourmodel{}, a framework designed to comprehensively calibrate TIR behaviors. We begin by analyzing common erroneous behavioral patterns in TIR models. Building on this analysis, we propose a pipeline that first enhances action space exploration and subsequently calibrates this exploration toward optimal trajectories. Empirical results demonstrate that \ourmodel{} exhibits effective behavioral patterns while ensuring inference correctness. We believe this work establishes a robust foundation for future research into TIR behaviors calibration.

\section*{Limitations}

Due to resource constraints, our experiments are restricted to local Wikipedia retrieval, and it is relatively difficult to apply the \ourmodel{} framework to larger models. In the future, we intend to scale the \ourmodel{} framework to larger architectures and integrate live web search capabilities, thereby broadening the scope of TIR behavioral exploration.





\bibliography{latex/custom}

\appendix


\section{Details about Datasets and Baselines}
\label{sec:appendix-dataset-baseline}
In this section, we introduce all the datasets and baselines used in the experiment.
\subsection{Datasets}
\label{app:dataset}
\subsubsection{Training Datasets}
Our training data is composed as shown in Table~\ref{tab:data_source}.
\paragraph{AIME}is derived from the American Invitational Mathematics Examination. The questions usually consist of 30 problems each year, covering Algebra, Geometry, Number Theory, Combinatorics, and Probability. It has become an important standard for measuring the reasoning level of models.
\paragraph{NuminaMATH}~\cite{numinamath} is developed by the Numina team and Hugging Face, aiming to significantly enhance LLMs' reasoning ability in solving mathematical problems. It contains approximately 860,000 pairs of mathematical problems and their detailed solutions, covering various difficulty levels from high school exercises to top international Olympic competitions.
\paragraph{OpenR1-Math-220k}~\cite{openr1} is a large-scale mathematical reasoning dataset released by the Open R1 team of Hugging Face in early 2025. It contains approximately 220k pieces of high-quality mathematical reasoning data. These data are directly generated locally using DeepSeek-R1 on 512 H100 graphics cards, covering various difficulty levels from basic education to competition level.

\begin{table}[t] 
    \caption{The composition of training data after sampling. For AIME, we use questions from 2023 and earlier. Abbreviation: MR (Mathematical-Reasoning Tasks), KI (Knowledge-Intensive Tasks), DS (Deepsearch Tasks).}
    \label{tab:data_source}
    \centering
    \small  
    \setlength{\tabcolsep}{2pt} 
    \begin{tabular}{lcc}
        \toprule
        \textbf{Dataset} & \textbf{Task} & \textbf{\# Counts} \\
        \midrule
        AIME & MR &  0.9k \\
        Numina-CoT~\cite{numinamath} & MR & 5k \\
        Numina-TIR~\cite{numinamath} & MR & 5k \\
        OpenR1-Math-220k~\cite{openr1} & MR & 10k \\
        DAPO-Math-17k~\cite{dapo} & MR & 5k\\
        \midrule
        2WikiMultiHopQA~\cite{2wiki} & KI & 5k\\
        HotpotQA~\cite{hotpotqa} & KI & 5k\\
        MuSiQue~\cite{musique} & KI & 5k\\
        TriviaQA~\cite{triviaqa} & KI & 5k\\
        \midrule
        ASearcherLRM35k~\cite{beyond10turns} & DS & 10k \\
        WebShaper~\cite{webshaper} & DS & 0.5k\\
        \bottomrule
    \end{tabular}
\end{table}

\paragraph{DAPO}~\cite{dapo} is jointly launched by ByteDance and Tsinghua University's AIR Institute, aiming to solve the training challenges of LLMs in Long-CoT. DAPO-Math-17k is the core open-source dataset of this project, specifically designed to enhance the mathematical reasoning ability of models.
\paragraph{2WikiMultiHopQA}~\cite{2wiki} is a benchmark dataset used to evaluate the multi-hop reasoning capabilities of LLMs. It is proposed in 2020 at the COLING conference, aiming to address issues present in mainstream datasets at the time (such as HotpotQA), including ``incomplete logical chains'' and ``solving problems through shortcuts without requiring multi-hop reasoning''.
\paragraph{HotpotQA}~\cite{hotpotqa} is a highly renowned large-scale multi-hop question-answering dataset in the field of Natural Language Processing (NLP). The original design intent is to advance machines' capabilities in human-like reasoning chains. Unlike traditional question-answering datasets, HotpotQA requires models to integrate information from multiple different documents and perform reasoning to arrive at an answer.
\paragraph{MuSiQue}~\cite{musique} is currently a highly challenging multi-hop question-answering benchmark in the field of Natural Language Processing. Instead of directly asking crowdsourcing personnel to write complex questions, it starts from simple ``single-hop questions'' and constructs complex 2-4 hop questions through rigorous logical combinations. It contains approximately 25,000 questions, covering logical depths of 2 to 4 hops.
\paragraph{TriviaQA}~\cite{triviaqa} is a highly classic and challenging large-scale open-domain question answering and reading comprehension dataset in the field of Natural Language Processing, primarily consisting of ``question-answer-evidence document'' triples. The original design intent of TriviaQA was to provide an evaluation benchmark that is closer to the real world and more difficult to solve through simple keyword matching.
\paragraph{ASearcherLRM35k}~\cite{beyond10turns} is a dataset autonomously generated by a Data Synthesis Agent. The core value of this dataset lies in addressing the pain points of data scarcity and overly short interaction turns for search agents during training. Through a fully automated synthesis workflow, it constructs a high-quality task library that supports models in performing ``long-range search reasoning''.
\paragraph{WebShaper}~\cite{webshaper} is a formal-driven information-seeking data synthesis framework and dataset released by Alibaba Tongyi Lab in 2025. WebShaper shifts toward a ``formal-driven'' approach, utilizing mathematical set theory to systematically construct complex search and reasoning tasks.

\subsubsection{Evaluation Datasets}
\paragraph{AMC23}is an important benchmark in the field of artificial intelligence used to evaluate the competition-level mathematical reasoning capabilities of LLMs. It primarily consists of real exam questions from the 2023 American Mathematics Competitions (AMC 10 and AMC 12). It can effectively test whether a model truly possesses reasoning abilities or is merely ``memorizing'' old problems from its training set.
\paragraph{MATH500}~\cite{math500} is currently one of the core benchmarks for measuring the competition-level mathematical reasoning capabilities of LLMs. Its emergence was primarily to address the issue of excessive evaluation costs associated with full-scale datasets, while ensuring that the evaluation problems possess sufficient discriminative power. It covers seven mathematical subfields, including Algebra, Geometry, and Number Theory.
\paragraph{Bamboogle}~\cite{bamboogle} is a challenging dataset specifically designed to evaluate the multi-hop reasoning and Retrieval-Augmented Generation (RAG) capabilities of LLMs. It is used to test whether search agents can piece together the final answer through multiple, multi-dimensional searches.
\subsection{Baselines}
\paragraph{Direct Inference}We use the Qwen2.5-7B-Instruct~\cite{qwen2.5} model to directly perform reasoning on the task set, and the prompt is shown in Figure~\ref{app:prompt_tir}.
\paragraph{Search-o1}~\cite{search-o1} is a framework designed to enhance the search reasoning capabilities of LLMs. Its core concept is to combine agentic search with slow thinking reasoning processes to solve the knowledge deficit or hallucination problems. The system triggers a search only when the model generates specific search instructions (such as query terms with special identifiers) within its thinking process.
\paragraph{Search-R1}~\cite{search-r1} is a novel model training framework that deeply integrates Reinforcement Learning with search engines. Search-R1 enables the model to learn like a human researcher: thinking while deciding whether a search is necessary, and refining its thought process based on search results through multiple iterations. This RL-based training approach allows the model to autonomously evolve highly efficient search strategies.
\paragraph{Research}~\cite{research} The core idea of research is to enable LLMs to autonomously learn, through reinforcement learning, when to search the web and how to utilize the search results, thereby solving complex multi-hop reasoning problems. By providing the model with search tools, it spontaneously learns search strategies through trial and error.
\paragraph{WebThinker}~\cite{webthinker} empowers large reasoning models with the ability to think, search, and write simultaneously like a human researcher. It breaks static limitations, granting the model the power to autonomously explore the internet. WebThinker performs excellently across multiple highly challenging benchmarks.
\paragraph{WebSailor}~\cite{websailor} is an advanced AI Agent training method proposed by Alibaba Tongyi Lab in 2025. Its core goal is to enable open-source large models to possess reasoning capabilities comparable to closed-source systems when handling complex web information retrieval tasks. Experimental results indicate that WebSailor achieves a qualitative leap compared to previous open-source agents when faced with extremely difficult tasks.
\paragraph{ToRL}~\cite{torl} is a framework allowing large models to autonomously learn how to use tools through trial and error, rather than imitating templates provided by humans. During the reasoning process, the model generates code, the system immediately invokes an interpreter to run it, and feeds the results back to the model. In high-difficulty mathematics competitions such as AIME24, the performance of the ToRL-7B model can surpass that of large-scale models without tools.
\paragraph{DotaMath}~\cite{dotamath} is a novel training and inference framework specifically designed to enhance the mathematical reasoning capabilities of LLMs. By introducing methods such as decomposition of thought, code assistance, and self-correction, DotaMath aims to address the issue of traditional LLMs being prone to errors when processing complex mathematics (such as the MATH dataset).
\paragraph{START}~\cite{start} is a brand-new training framework for LLMs proposed by the Alibaba research team in March 2025. This method constructs training data to fine-tune models through prompt-guided reasoning and Hint Rejection Sampling Fine-Tuning. The START model, trained based on the QwQ-32B base model, has delivered astonishing performance on multiple hardcore leaderboards.
\paragraph{IKEA}~\cite{ikea} is a brand-new method aimed at improving the efficiency of AI Search Agents. This method is designed to address the current issues where AI agents either over-rely on external searches or ignore their own internal knowledge when processing tasks. The IKEA method exhibits advantages such as high efficiency, greater accuracy, and strong generalization when handling knowledge-intensive tasks.
\paragraph{SMART}~\cite{smart} is a research framework designed to address the tool-overuse problem that occurs when LLM agents invoke external tools. The SMART method is inspired by human Metacognition, or ``thinking about thinking''. It aims to calibrate the agent's self-awareness, enabling it to determine when to use its own knowledge to think and when to use tools to search.
\paragraph{AutoTIR}~\cite{autotir} is a novel post-training framework for LLMs. The core purpose of this method is to allow the model to autonomously decide when to use a tool and which tool to use during the reasoning process, rather than following predefined, rigid templates. AutoTIR achieves an excellent balance between core language capabilities and tool integration capabilities.
\paragraph{Tool-Star}~\cite{dong2025toolstarempoweringllmbrainedmultitool} is a Reinforcement Learning based framework designed to enhance the ability of LLMs to use multiple external tools autonomously, efficiently, and collaboratively during complex reasoning processes. Through two-stage training and systematic data synthesis, Tool-Star aims to solve problems such as the difficulty of multi-tool collaboration and low tool-calling efficiency.



\section{Group-wise Pareto Sampling Flow}
\label{app:pareto sampling}
In this section, we introduce the algorithmic process of Group-wise Pareto Sampling, as shown in Algorithm~\ref{alg:pareto sampling} and~\ref{alg:computedistance}.

\begin{algorithm}[H]
\small
\caption{Group-wise Pareto Sampling}
\textbf{Input} Dataset $D_{\text{source}}$, target selection size $N$, objectives $S_{\text{corr}}$ and $S_{\text{tool}}$
\begin{algorithmic}[1]
\For{$x$ in $D_{\text{source}}$} 
    \State Calculate dispersion metrics $S_{\text{corr}}(x)$ and $S_{\text{tool}}(x)$ based on $K$ sampled trajectories
\EndFor 
\State Partition $D_{\text{source}}$ into hierarchical Pareto frontiers $\{\mathcal{F}_1, \mathcal{F}_2, \dots, \mathcal{F}_k\}$ according to relation $\succ$
\State Initialize $S \leftarrow \emptyset$
\For{$\mathcal{F}$ in $\{\mathcal{F}_1, \mathcal{F}_2, \dots, \mathcal{F}_k\}$} 
    \If{$|\mathcal{S}| + |\mathcal{F}| \le N$}
        \State $\mathcal{S} \leftarrow \mathcal{S} \cup \mathcal{F}$
    \Else
        \State $\text{Distances} \leftarrow \text{ComputeCrowdingDistance}(\mathcal{F})$
        \State Sort $\mathcal{F}$ by $\text{Distances}$ in descending order
        \State $\mathcal{S} \leftarrow \mathcal{S} \cup \text{Top}(N - |\mathcal{S}|)$ samples from $\mathcal{F}$
    \EndIf
\EndFor
\end{algorithmic}
\textbf{Output} $\mathcal{S}$
\label{alg:pareto sampling}
\end{algorithm}

\begin{algorithm}[H]
\small
\caption{ComputeCrowdingDistance}
\textbf{Input} $\mathcal{F}$
\begin{algorithmic}[1]
\State Initialize Distance $D$
\For{$j$ in \text{range}$(D)$} 
    \State Initialize $D_j \leftarrow 0$
\EndFor 
\For{\text{objective} $m$ in \{$S_{\text{corr}}, S_{\text{tool}}$\}} 
    \State Sort $\mathcal{F}$ by objective $m$ in ascending order
    \For{$j$ in \text{range}$(\mathcal{F})$}
        \State $D_j \leftarrow D_j + \frac{m(j+1) - m(j-1)}{\text{Range}(m)}$
    \EndFor
\EndFor
\end{algorithmic}
\textbf{Output} $D$
\label{alg:computedistance}
\end{algorithm}

\section{Implementation Details}
\label{app:implementation details}
\subsection{Preliminary Experiments Details}
\label{app:pre exp details}
For preliminary experiments, mathematical task includes AIME, Numina-Math, and OpenR1-Math-220k; knowledge-intensive task includes 2WikiMultihopQA, HotpotQA, MuSiQue, and TriviaQA. For insufficient tool call analysis experiment, approximately 20.8\% of failures can be corrected after continued reasoning. Hints used for injection can be found in Figure~\ref{fig:hints}. For the analysis experiment on the tool call number differences, we sample 16 trajectories for each query. 
\subsection{Training Details}
\paragraph{Details in Action Space Exploration} After data sampling, we select a total of 10.5k data for RFT. Our base model uses Qwen2.5-7B-Instruct~\cite{qwen2.5}, with a learning rate of 5e-6 and a batch size set to 32. We train for a total of three epochs. During training, we use DeepSpeed ZeRO-3~\cite{deepspeed} and FlashAttention2~\cite{flashattn}.
\paragraph{Details in RL Training} We use the VERL framework~\cite{verl} and train based on the ARPO algorithm~\cite{arpo}. We conduct three rounds of curriculum learning. The batch size is 64, we train for 3 epochs, the maximum length of the output text is 4096 tokens, and the learning rate is 1e-6. The initial values of both types of sigma are set to 0.1. All experiments are completed using 4 NVIDIA A800 GPUs. During training, we refer to FlashRAG~\cite{flashrag} and use Wikipedia documents for local retrieval. For each question, we retrieve 4 documents, and each document has a maximum length of 1200 characters.
\subsection{Evaluation Details}
For each test dataset, we select the first 500 samples for testing. For Mathematical-Reasoning Tasks, we use Google Search. For Knowledge-Intensive Tasks, we use local retrieval consistent with the training settings. We set the maximum call number for both the code compiler and the search engine to 6 times each. The maximum output sequence length for each round of reasoning is set to 4096 tokens.

\begin{table}[t]
\small
\centering
\setlength{\tabcolsep}{0.5mm} 
\begin{tabular}{lccccccc}
\toprule
 \textbf{Method} & AIME24 & AMC23 & M500 & 2Wiki & Bamb & MSQ & Avg. \\
\midrule
IKEA & \underline{33.3} & 30.0 & 24.4 & 30.0 & 28.8 & \textbf{31.5} & 29.7\\
SMART & 6.7 & 20.0 & 33.6 & 19.0 & 33.6 & 12.0 & 20.8\\
AutoTIR & 23.3 & 50.0 & 60.4 & 47.5 & \underline{53.6} & \underline{31.0} & 44.3\\
Tool-Star & 30.0 & 55.0 & 68.2 & \textbf{64.5} & 47.2 & 22.0 & \underline{47.8}\\
Tool-Light & 32.5 & \textbf{72.5} & \underline{75.8} & 32.5 & 25.6 & 13.5 & 42.1\\
\ourmodel{} & \textbf{53.3} & \underline{65.0} & \textbf{84.6} & \underline{62.0} & \textbf{58.4} & 18.5 & \textbf{57.0}\\
\bottomrule
\end{tabular}
\caption{
Results of \textbf{Conciseness} metric, with top two results highlighted in \textbf{bold} and \underline{underlined}. Abbreviation: M500 (MATH500).}
\label{tab:conciseness}
\end{table}

\begin{table}[t]
\small
\centering
\setlength{\tabcolsep}{0.5mm} 
\begin{tabular}{lccccccc}
\toprule
 \textbf{Method} & AIME24 & AMC23 & M500 & 2Wiki & Bamb & MSQ & Avg. \\
\midrule
IKEA & 10.0 & 42.5 & 63.2 & 33.8 & 35.0 & 14.2 & 33.1\\
SMART & 3.3 & 17.5 & 39.1 & 21.6 & 40.8 & 12.1 & 22.4\\
AutoTIR & 26.4 & 56.5 & 69.0 & \underline{64.6} & \textbf{56.9} & \textbf{30.7} & 50.7\\
Tool-Star & 30.0 & \underline{57.5} & \underline{79.4} & \textbf{65.8} & \underline{55.3} & 24.4 & \underline{52.1}\\
Tool-Light & \underline{33.3} & \textbf{70.0} & 75.6 & 60.2 & 48.1 & \underline{25.5} & 52.1\\
\ourmodel{} & \textbf{43.3} & \textbf{70.0} & \textbf{84.5} & 63.2 & 54.3 & 23.3 & \textbf{56.4}\\
\bottomrule
\end{tabular}
\caption{
Results of \textbf{Successful Execution} metric, with top two results highlighted in \textbf{bold} and \underline{underlined}. Abbreviation: M500 (MATH500).}
\label{tab:success exe}
\end{table}

\begin{table}
\small
\centering
\setlength{\tabcolsep}{0.5mm} 
\begin{tabular}{lccccccc}
\toprule
 \textbf{Method} & AIME24 & AMC23 & M500 & 2Wiki & Bamb & MSQ & Avg. \\
\midrule
IKEA & 0.17 & 0.75 & 2.19 & 1.49 & 2.10 & 0.47 & 1.20\\
SMART & 0.13 & 0.33 & 1.07 & 1.20 & 0.98 & 0.25 & 0.66\\
AutoTIR & \textbf{1.00} & \textbf{2.32} & \textbf{3.60} & \underline{3.61} & \underline{3.85} & \underline{1.49} & \underline{2.65}\\
Tool-Star & 0.42 & 0.76 & 1.56 & 2.24 & 1.34 & 0.57 & 1.15\\
Tool-Light & 0.47 & 1.19 & 1.86 & 1.96 & 1.80 & 0.78 & 1.34\\
\ourmodel{} & \underline{0.59} & \underline{1.23} & \underline{2.84} & \textbf{6.34} & \textbf{5.62} & \textbf{1.81} & \textbf{3.07}\\
\bottomrule
\end{tabular}
\caption{
Results of \textbf{Reasoning Length} metric, with top two results highlighted in \textbf{bold} and \underline{underlined}. Abbreviation: M500 (MATH500).}
\label{tab:length}
\end{table}

\section{Complete Results of Behavioral Pattern Indicators}
\label{app:full results of radar}

In this section, we present the complete results on all benchmarks shown in Figure~\ref{fig:radar}. As depicted in Table~\ref{tab:conciseness},~\ref{tab:success exe}, and~\ref{tab:length}, \ourmodel{} outperforms most baseline methods on all metrics, and stably surpasses them in terms of average performance. This highlights the effectiveness of \ourmodel{}'s training process.

\section{Prompt Templates}
\label{app:prompt}
\subsection{Prompt for Redundancy Judgment}
\label{app:prompt_redundancy}
    
\begin{tcolorbox}[
    colframe = gray,       
    colback = gray!5!white,             
    coltitle = white,                   
    coltext = black,                    
    fonttitle = \bfseries,              
    title = Prompt for Redundancy Judgment,  
    boxrule = 1pt,                      
    arc = 2mm,                          
    width = \linewidth,                 
    left = 7pt,                         
    right = 7pt,                        
    top = 5pt,                          
    bottom = 5pt                        
]
\fontsize{8.5pt}{10pt}\selectfont
You are an expert analyst of tool-integrated reasoning tasks. 
Given a question, its golden answer, and a reasoning trajectory, your task is to examine the following content and determine whether the trajectory contains any redundant or ineffective tool calls.\\

[Question]\\
\{question\}

[Golden Answer]\\
\{answer\}

[Trajectory]\\
\{trajectory\}

Definitions:\\
- Redundant tool call: A call whose result is repetitive of previous content, unused in later reasoning, or does not contribute to the final answer.\\
- Ineffective tool call: A call that returns an irrelevant or unhelpful result providing no meaningful contribution to the reasoning process.\\

Your tasks:\\
1. Determine whether any redundant or ineffective tool calls exist in the trajectory.\\
2. If such calls exist, output only ``<answer>yes</answer>''.\\
3. If none exist, output only ``<answer>no</answer>''.\\

Output format:\\
Only output ``yes'' or ``no'', enclosed within <answer> and </answer> tags, and do not output anything else.
\end{tcolorbox}

\begin{figure}[t]
  \includegraphics[width=\columnwidth]{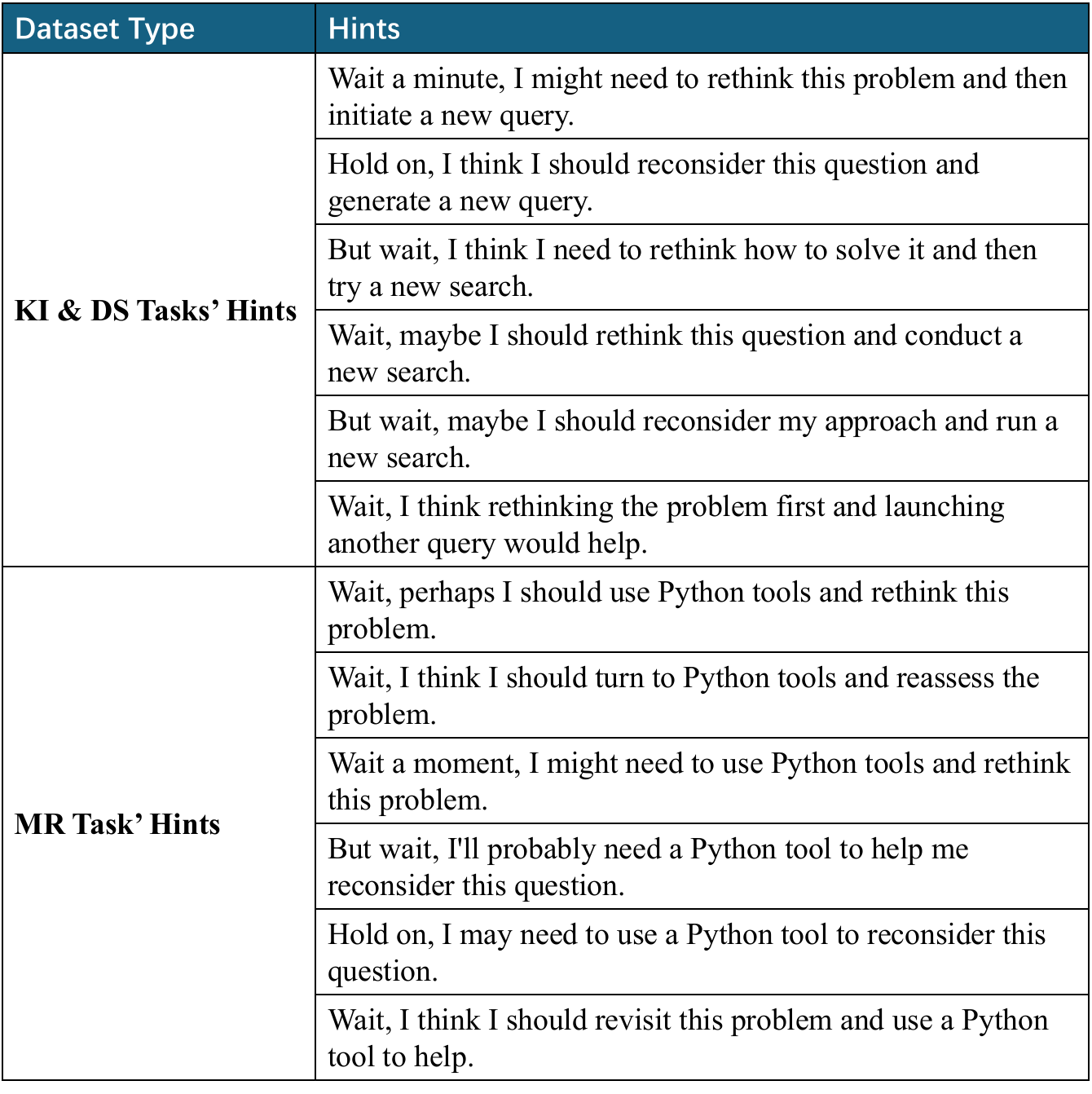}
  \caption{All hints used in Incorrect Trajectories Reflection stage.}
  \label{fig:hints}
\end{figure}

\subsection{Prompt for TIR Trajectory Sampling}
\label{app:prompt_tir}
\begin{tcolorbox}[
    colframe = gray,       
    colback = gray!5!white,             
    coltitle = white,                   
    coltext = black,                    
    fonttitle = \bfseries,              
    title = Prompt for TIR Trajectory Sampling,  
    boxrule = 1pt,                      
    arc = 2mm,                          
    width = \linewidth,                 
    left = 7pt,                         
    right = 7pt,                        
    top = 5pt,                          
    bottom = 5pt                        
]
\fontsize{8.5pt}{10pt}\selectfont
You are a helpful assistant that can solve the given question step by step with the help of the wikipedia search tool and python interpreter tool. \
Given a question, you need to first think about the reasoning process in the mind and then provide the answer. \
During thinking, you can invoke the wikipedia search tool to search and python interpreter tool to calculate the math problem for fact information about specific topics if needed. \
The reasoning process and answer are enclosed within <think> </think> and <answer> </answer> tags respectively, \
and the search query and result are enclosed within <search> </search> and <result> </result> tags respectively. \
After receiving the search or python result, you should continue your reasoning process begin with <think>. \
For example, <think> This is the reasoning process. </think> <search> search query here </search> <result> search result here </result> \
<think> This is the reasoning process. </think> <python> python code here </python> <result> python interpreter result here </result> \
<think> This is the reasoning process. </think> <answer> The final answer is \[ \boxed{answer\ here} \] </answer>. \
In the last part of the answer, the final exact answer is enclosed within \textbackslash boxed\{\} with latex format.\footnote{\scriptsize Unless otherwise specified, we use this prompt for reasoning during evaluation.}
\end{tcolorbox}

\subsection{Prompt for First Step Optimization}
\label{app:prompt_correct_first_redundancy_find}
\begin{tcolorbox}[
    colframe = gray,       
    colback = gray!5!white,             
    coltitle = white,                   
    coltext = black,                    
    fonttitle = \bfseries,              
    title = Prompt for First Step Optimization,  
    boxrule = 1pt,                      
    arc = 2mm,                          
    width = \linewidth,                 
    left = 7pt,                         
    right = 7pt,                        
    top = 5pt,                          
    bottom = 5pt                        
]
\fontsize{8.5pt}{10pt}\selectfont
\#\# Role:\\
You are an expert in analyzing tool-integrated reasoning trajectory. \\

\#\# Task Description:\\
Your goal is to analyze a given trajectory to identify and locate the first occurrence with any redundant reasoning or redundant tool usage. If any redundancy is identified, you should find the earliest step that exists redundancy, and provide an analysis and a simplified, more efficient version for that step.\\

\#\# Redundancy Definitions:\\
1. Redundant Reasoning: The thought process includes unnecessary, overly complex, verbose, or repetitive content.\\
2. Redundant Tool Call: A tool is called when the necessary information is already sufficient, or the tool call's content is highly repetitive of a prior one seeking no new information.\\
  
\#\# Input Data:\\
\#\#\# Question: \{question\}\\
\#\#\# Trajectory: \\
\{trajectory\}\\

\#\# Output Format:\\
Your output must be one of the following two formats, with no additional explanations or text.\\
\textbf{Scenario 1: The trajectory does not have any redundant reasoning or redundant tool usage.}
Only output ``Step: no'', and do not output anything else.\\
\textbf{Scenario 2: The trajectory exists any redundant reasoning or redundant tool usage.}
Your output must be exactly three lines. The first line is the earliest step number with redundancy, the second line is the analysis of the redundancy in that step, and the third line is a simplified, more efficient version of that specific step. In this scenario, the output format is as follows:\\
Step: [an integer step number]\\
Analysis: [A concise analysis of the redundancy in that step]\\
Corrected Step: [the full, simplified step string]\\

For example:\\
In scenario 1, the output is ``Step: no''.\\

In scenario 2, the output format is as follows:\\
Step: [an integer step number]\\
Analysis: [A concise analysis of the redundancy in that step]\\
Corrected Step: <think> This is the reasoning process. </think> <search> search query here </search> <result> search result here </result> <think> This is the reasoning process. </think> <python> python code here </python> <result> python interpreter result here </result> <think> This is the reasoning process. </think> <answer> The final answer is \[ \boxed{answer\ here} \] </answer>.
\end{tcolorbox}

\subsection{Prompt for Global Refinement}
\label{app:prompt_global_refinement}
\begin{tcolorbox}[
    colframe = gray,       
    colback = gray!5!white,             
    coltitle = white,                   
    coltext = black,                    
    fonttitle = \bfseries,              
    title = Prompt for Global Refinement,  
    boxrule = 1pt,                      
    arc = 2mm,                          
    width = \linewidth,                 
    left = 7pt,                         
    right = 7pt,                        
    top = 5pt,                          
    bottom = 5pt                        
]
\fontsize{8.5pt}{10pt}\selectfont
You are an expert in trajectory refinement. You will be given a question and a corresponding trajectory that uses a fixed format with tags such as ``<think>'', ``<search>'', ``<result>'', ``<python>'', and ``<answer>''.\\

Your task is to \textbf{refine ONLY the text inside ``<think>...</think>'' tags}, while keeping the trajectory's structure, tag order, and all other content unchanged.\\

\#\#\# Requirements:\\

1. \textbf{Do NOT add, remove, reorder, or rename any tags.}\\
   The sequence of tags must remain exactly the same as in the original trajectory.
   For example, if the format is:\\
   ``<think>...</think> <search>...</search> <result>...</result> <think>...</think> <answer>...</answer>''\\
   then the refined version must follow the exact same pattern.\\

2. \textbf{Only modify the content inside ``<think>'' tags.}\\

   * Remove redundant thoughts or irrelevant reasoning.\\
   * Eliminate obvious mistakes or unnecessary repetition.\\
   * Preserve the core reasoning and logical flow.\\
   * Keep the refined version concise, clear, and logically consistent.\\

3. \textbf{Do NOT alter anything inside ``<search>'', ``<result>'', ``<python>'', or ``<answer>'' tags.}\\
   Their content, formatting, and text must remain identical.\\

4. \textbf{Keep the formatting and tags unchanged.}\\

---\\

\#\#\# Input\\

\textbf{[Question]}\\
\{question\}\\

\textbf{[Original Trajectory]}\\
\{trajectory\}\\

---\\

\#\#\# Output\\

Provide the \textbf{refined trajectory}, keeping the same structure and identical content outside the ``<think>'' tags, with only streamlined and corrected content inside each ``<think>'' section.\\

Your output:
\end{tcolorbox}

\subsection{Prompt for Self Correction}
\label{app:prompt_self_correction}
\subsubsection{Prompt for First Flaw Identification}
\begin{tcolorbox}[
    colframe = gray,       
    colback = gray!5!white,             
    coltitle = white,                   
    coltext = black,                    
    fonttitle = \bfseries,              
    title = Prompt for First Flaw Identification,  
    boxrule = 1pt,                      
    arc = 2mm,                          
    width = \linewidth,                 
    left = 7pt,                         
    right = 7pt,                        
    top = 5pt,                          
    bottom = 5pt                        
]
\fontsize{8.5pt}{10pt}\selectfont
\#\# Role\\
You are an expert in analyzing tool-integrated reasoning trajectories.\\

\#\# Task Description\\
You will be given a reasoning trajectory that was produced by you in a previous turn. Your goal is to examine this trajectory step by step and identify the \textbf{first} step that contains any reasoning flaw or improper tool call. Once located, provide your analysis and practical suggestions for that step.\\

\#\# Flaw Categories\\
1. Flawed Reasoning:\\
   - Logical Error: The thought process is incorrect, irrelevant, or does not contribute to solving the goal.\\
   - Overthinking: The reasoning is unnecessarily long, repetitive, or contains redundant content.\\

2. Improper Tool Call:\\
   - Redundant Call: A tool is called even though the available information is already sufficient, or the call unnecessarily repeats prior content.\\
   - Inappropriate Call: The selected tool is unsuitable for the task, or the content passed to the tool is inappropriate.\\

\#\# Input\\
\#\#\# Question:\\
\{question\}\\

\#\#\# Trajectory:\\
\{trajectory\}\\

\#\# Output Requirements\\
Your response must contain \textbf{exactly two lines}, with no additional explanation, formatting, or text:\\

Line 1:\\
Step: [the integer index of the earliest flawed step]\\

Line 2:\\
Analysis: [a concise analysis and practical suggestions of the flaw in that step]\\

Among them, your analysis and practical suggestions should be written in a \textbf{single paragraph of plain text}. Do not use any headings, bullet points, numbering, or formatting. Just give a concise summary of the main flaws you found in the trajectory and how it could be improved, in continuous natural language.\\

For example:\\
Step: [an integer step number] \\
Analysis: [A paragraph with analysis and practical suggestions of the flaw]\\

Your output:
\end{tcolorbox}

\subsubsection{Prompt for First Flaw Modification}
\begin{tcolorbox}[
    colframe = gray,       
    colback = gray!5!white,             
    coltitle = white,                   
    coltext = black,                    
    fonttitle = \bfseries,              
    title = Prompt for First Flaw Modification,  
    boxrule = 1pt,                      
    arc = 2mm,                          
    width = \linewidth,                 
    left = 7pt,                         
    right = 7pt,                        
    top = 5pt,                          
    bottom = 5pt                        
]
\fontsize{8.5pt}{10pt}\selectfont
You are an expert in refining and completing tool-integrated reasoning trajectories.\\

\#\# Task Description:\\
You will be given:\\
1. A question\\
2. An incomplete trajectory\\
3. An analysis specifically about the \textbf{last step} in the trajectory\\

Your goal is to generate a \textbf{new, corrected final step} to replace the current last step based on the provided analysis. Only generate a new final step, do not generate any other content.\\

\#\# Input Data:\\
\#\#\# Question: \{question\}\\
\#\#\# Trajectory (incomplete):\\
\{trajectory\}\\
\#\#\# Analysis of the last step:\\
\{analysis\}\\

\#\# Output Format:\\
Your output must be exactly one line containing only the modified final step, in full.\\

For example:\\
your reasoning process. <search> search query here </search> <result> search result here </result> your reasoning process. <python> python code here </python> <result> python interpreter result here </result> your reasoning process. <answer> The final answer is \[ \boxed{answer\ here} \] </answer>.\\

Your modified step: 
\end{tcolorbox}

\subsection{Hints for Trajectory Reflection}
For mathematical-reasoning tasks and knowledge-intensive tasks, we design different hints respectively. They can be found in Table~\ref{fig:hints}.

\section{Case Study of Error Behavior Patterns}
\label{app:case study}
We provide a representative example for each of the four error patterns. They can illustrate the specific failure modes of the TIR model, as detailed in Table~\ref{tab:case_redundant},~\ref{tab:case_exe_error},~\ref{tab:case_insufficient} and~\ref{tab:case_reasoning}. \colorbox{myred}{Red} contents represent the erroneous places.

\begin{table*}[!t]
    \centering
    \caption{An example of Redundant Tool Call situation.}
    \fontsize{9pt}{11pt}\selectfont
    \renewcommand{\arraystretch}{0.8} 
    \begin{tabular}{p{0.95\linewidth}}
    \midrule
        \rowcolor{gray!20}\textbf{Example \#1 from MuSiQue Dataset} \\
    \midrule
        \textbf{Question:} \\
        What river does the system containing Umchabezi River serve as the mouth of?\\
        \textbf{Golden Answer:} \\
        Limpopo River \\
    \midrule
        \rowcolor{gray!20}
        \textbf{TIR Stragety}\\
    \midrule
        $<$think$>$ To find out which river system the Umchabezi River serves as the mouth of, I'll need to search for information on this river.
 $<$/think$>$\\$<$search$>$
Wikipedia search query: Umchabezi River mouth
$<$/search$>$\\$<$result$>$
The Umchabezi River is a tributary of the Mzingwane River in Beitbridge District and Gwanda District, Zimbabwe. The main dam on the river is Makado Dam, which supplies water for commercial irrigation.
$<$/result$>$\\$<$search$>$
Search for Mzingwane River mouth
$<$/search$>$\\
$<$result$>$
The Mzingwane River, formerly known Umzingwane River as or Umzingwani River is a major left-bank tributary of the Limpopo River in Zimbabwe. It rises near Fort Usher, Matobo District, south of Bulawayo and flows into the Limpopo River near Beitbridge, downstream of the mouth of the Shashe River and upstream of the mouth of the Bubye River.  The Mzingwane River is an ephemeral river with flow generally restricted to the months when rain takes place (November to March), with most flow recorded between December and February, except where it has been modified by dam operations.
$<$/result$>$\\
$<$think$>$ According to the search results, the river system that the Umchabezi River serves as a tributary of is the Limpopo River. \colorbox{myred}{Let's verify this with Python code.}$<$/think$>$\\$<$python$>$
\begin{lstlisting}[style=python]
print('Limpopo River')
\end{lstlisting}
$<$/python$>$\\$<$result$>$
Limpopo River
$<$/result$>$\\$<$answer$>$
Therefore, the final answer is \boxed{\text{Limpopo River}}.
$<$/answer$>$\\
    \bottomrule
    \end{tabular}
    \label{tab:case_redundant}
\end{table*}

\begin{table*}[!t]
    \centering
    \caption{An example of Aborted Tool Execution situation.}
    \fontsize{9pt}{11pt}\selectfont
    \renewcommand{\arraystretch}{0.8} 
    \begin{tabular}{p{0.95\linewidth}}
    \midrule
        \rowcolor{gray!20}\textbf{Example \#2 from MATH500 Dataset} \\
    \midrule
        \textbf{Question:} \\
        Let $T=\frac{7}{8}$. The number $T$ can be expressed as a reduced fraction $\frac{m}{n}$, where $m$ and $n$ are positive integers whose greatest common divisor is $1$ . The equation has two distinct real solutions. Compute the lesser of these two solutions.\\
        \textbf{Golden Answer:} \\
        $-8$ \\
    \midrule
        \rowcolor{gray!20}
        \textbf{TIR Stragety}\\
    \midrule
        $<$think$>$ We are given:\\- $ T = \frac{7}{8} $\\- We express $ T $ as a reduced fraction $ \frac{m}{n} $, so $ m = 7 $, $ n = 8 $\\- Then we consider the quadratic equation:    $  x^2 + (m + n)x + mn = 0  $Substitute $ m = 7 $, $ n = 8 $:- $ m + n = 7 + 8 = 15 $- $ mn = 7 \cdot 8 = 56 $\\So the equation becomes:$x^2 + 15x + 56 = 0$\\We are told this equation has two distinct real solutions, which it does because the discriminant is positive.\\The discriminant $ D $ of $ ax^2 + bx + c = 0 $ is:$D = b^2 - 4ac$\\Here, $ a = 1 $, $ b = 15 $, $ c = 56 $:$D = 15^2 - 4(1)(56) = 225 - 224 = 1 > 0$\\Yes — two distinct real roots.\\Now compute the roots using the quadratic formula:$x = \frac{-b \pm \sqrt{D}}{2a} = \frac{-15 \pm \sqrt{1}}{2}= \frac{-15 \pm 1}{2}$\\Thus,- One root: $ \frac{-15 + 1}{2} = \frac{-14}{2} = -7 $- Other root: $ \frac{-15 - 1}{2} = \frac{-16}{2} = -8 $\\These are equal? Wait! That’s not right!\\Wait — let's double-check:$\frac{-15 \pm 1}{2}$\\That gives:- $ (-15 + 1)/2 = -14/2 = -7 $- $ (-15 - 1)/2 = -16/2 = -8 $\\So the two roots are $ -7 $ and $ -8 $. So they are distinct.\\But now — what is the lesser of the two solutions?Clearly, $ -8 < -7 $, so the lesser solution is $ -8 $.\\But wait — let me confirm everything again carefully. Given:- $ T = \frac{7}{8} $- So $ m = 7 $, $ n = 8 $- Form the equation: $ x^2 + (m+n)x + mn = 0 $- Which is $ x^2 + 15x + 56 = 0 $- Discriminant: $ 15^2 - 4(1)(56) = 225 - 224 = 1 $- Roots: $ \frac{-15 \pm 1}{2} $- So roots: $ \frac{-14}{2} = -7 $, $ \frac{-16}{2} = -8 $- Yes, both valid; $ -8 < -7 $\\Therefore, the lesser solution is $ -8 $. Let's use Python to check this answer.
 $<$/think$>$$<$python$>$
\begin{lstlisting}[style=python]
from sympy import symbols, Eq, solve

m, n = 7, 8
expr = symbols('x')
equation = Eq(x**2 + (m+n)*x + m*n, 0)
roots = solve(equation, x)
min_root = min(roots)
min_root.evalf()
\end{lstlisting}
$<$/python$>$$<$result$>$
\colorbox{myred}{Traceback (most recent call last):  File $<$string$>$, line 4, in $<$module$>$ }

\colorbox{myred}{NameError: name `x' is not defined}
$<$think$>$ Oops — I forgot to define $x$. Let me fix.
 $<$/think$>$
 $<$python$>$
\begin{lstlisting}[style=python]
from sympy import symbols, Eq, solve
x = symbols('x') 
equation = Eq(x**2 + (7+8)*x + 7*8, 0) 
roots = solve(equation, x) 
min_root = min(roots) 
print(f"{min_root.evalf()=}")
\end{lstlisting}
$<$/python$>$
$<$result$>$
min\_root.evalf()=-8.00000000000000
$<$/result$>$
$<$think$>$Hence, the lesser solution is:$<$/think$>$$<$answer$>$
\boxed{-8}.
$<$/answer$>$\\
    \bottomrule
    \end{tabular}
    \label{tab:case_exe_error}
\end{table*}

\begin{table*}[!t]
    \centering
    \caption{An example of Insufficient Tool Call situation.}
    \fontsize{9pt}{11pt}\selectfont
    \renewcommand{\arraystretch}{0.8} 
    \begin{tabular}{p{0.95\linewidth}}
    \midrule
        \rowcolor{gray!20}\textbf{Example \#3 from 2WikiMultiHopQA Dataset} \\
    \midrule
        \textbf{Question:} \\
        Who is Sancho Ramírez's maternal grandfather?\\
        \textbf{Golden Answer:} \\
        Bernard-Roger, Count of Bigorre \\
    \midrule
        \rowcolor{gray!20}
        \textbf{TIR Stragety}\\
    \midrule
        $<$think$>$ To determine who Sancho Ramírez's maternal grandfather was, we would need to know some biographical details about him, specifically focusing on his mother's side of the family. However, without access to specific historical records or more context about which Sancho Ramírez is being referred to, it's challenging to provide an accurate answer directly. Let's assume we're looking for information based on a notable figure named Sancho Ramírez. I will proceed by searching for general information on famous figures with this name.
 $<$/think$>$\\$<$search$>$
Sancho Ramírez biography
$<$/search$>$\\$<$result$>$
Sancho Ramírez was an illegitimate son of King Ramiro I of Aragon and Amua, the firstborn and brother of his namesake who would inherit the throne and reign as Sancho Ramírez.  Even though he could not inherit the throne because his father had legitimate issue, he was named count at an early age and was a prominent member of the \"curia regis\" first appearing in a charter dated 1049, suspected of being false, as a witness to a donation made by his father to the Monastery of San Victorin. In this charter, he confirms as \"Sancius Ranimiri regis filius primogenitus\" (Sancho, the firstborn son of King Ramiro) followed by his brother and namesake, Sancho, who confirms as \"Sancius Ranimiri regis filius prolis Ermmisendis regine\" (Sancho son of King Ramiro and Queen Ermesinda).
$<$/result$>$\\$<$think$>$The provided summary doesn’t contain explicit information about Sancho Ramírez’s maternal grandfather. \colorbox{myred}{Further targeted searches may be required to find this particular detail. Based on the available data, there isn't enough} \colorbox{myred}{evidence to identify Sancho Ramírez's maternal grandfather clearly.}$<$/think$>$\\$<$answer$>$
\boxed{\text{Insufficient information}}.
$<$/answer$>$\\
    \bottomrule
    \end{tabular}
    \label{tab:case_insufficient}
\end{table*}

\begin{table*}[!t]
    \centering
    \caption{An example of Erroneous Reasoning Process situation.}
    \fontsize{9pt}{11pt}\selectfont
    \renewcommand{\arraystretch}{0.8} 
    \begin{tabular}{p{0.95\linewidth}}
    \midrule
        \rowcolor{gray!20}\textbf{Example \#4 from MATH500 Dataset} \\
    \midrule
        \textbf{Question:} \\
        What is the smallest positive perfect cube that can be written as the sum of three consecutive integers?\\
        \textbf{Golden Answer:} \\
        $27$ \\
    \midrule
        \rowcolor{gray!20}
        \textbf{TIR Stragety}\\
    \midrule
        $<$think$>$ To find the smallest positive perfect cube that can be written as the sum of three consecutive integers, let's start by expressing the sum of three consecutive integers: $n-1$, $n$, and $n+1$ which equals $3n$. So we're looking for the smallest positive perfect cube that can be expressed as $3n$.

A perfect cube is a number of the form $k^3$ where $k$ is an integer. We need $k^3 = 3n$, implying that $k^3$ must be divisible by 3.

\colorbox{myred}{Let's check the smallest values for $k$ such that $k^3$ is a multiple of 3.}
 $<$/think$>$\\$<$search$>$
smallest perfect cubes that are multiples of 3
$<$/search$>$\\$<$result$>$
Perfect magic cube  In mathematics, a perfect magic cube is a magic cube in which not only the columns, rows, pillars, and main space diagonals, but also the cross section diagonals sum up to the cube's magic constant.  Perfect magic cubes of order one are trivial; cubes of orders two to four can be proven not to exist, and cubes of orders five and six were first discovered by Walter Trump and Christian Boyer on November 13 and September 1, 2003, respectively. A perfect magic cube of order seven was given by A. H. Frost in 1866, and on March 11, 1875, an article was published in the Cincinnati Commercial newspaper on the discovery of a perfect magic cube of order 8 by Gustavus Frankenstein.$<$/result$>$\\$<$think$>$It seems that the smallest value of $k$ cannot be found through searching. Let's try using Python.$<$/think$>$\\
$<$python$>$
\begin{lstlisting}[style=python]
# Check the smallest k that makes k^3 a multiple of 3
for k in range(1, 10):
    if (k**3) % 3 == 0:
        smallest_cube_multiple_of_3 = k**3
        break

smallest_cube_multiple_of_3
\end{lstlisting}
$<$/python$>$\\
$<$result$>$
$27$
$<$/result$>$\\
$<$think$>$From our calculations and understanding, the smallest positive perfect cube that can be written as the sum of three consecutive integers is $27$. Since $27 = 3^3$ and it fits the requirement as $3 \times 9$, it is indeed the correct answer.$<$/think$>$\\
$<$answer$>$
The smallest positive perfect cube that can be written as the sum of three consecutive integers is \boxed{27}. 
$<$/answer$>$\\
    \bottomrule
    \end{tabular}
    \label{tab:case_reasoning}
\end{table*}

\end{document}